\documentclass[11pt,draftcls,onecolumn]{IEEEtran}

\usepackage{amsmath,amsfonts,amssymb,amsbsy}
\usepackage{float}
\usepackage{verbatim}
\usepackage{pslatex}
\usepackage{cite,url,bm}
\usepackage{epsf,subfigure,psfig}
\usepackage{latexsym,amsmath,epsfig}
\usepackage{graphicx}
\usepackage{algorithmic}
\usepackage{algorithm}

\newcommand{\vect}[1]{\pmb{#1}}
\newcommand{\mat}[1]{\pmb{#1}}

\newcommand{\norm}[1]{\left\|#1\right\|}
\newcommand{\R}{\mathbb{R}}

\def\mb{\mathbf}

\def\ben{\begin{equation*}}
\def\een{\end{equation*}}
\def\be{\begin{equation}}
\def\ee{\end{equation}}
\def\beaa{\begin{eqnarray*}}
\def\eeaa{\end{eqnarray*}}
\def\bea{\begin{eqnarray}}
\def\eea{\end{eqnarray}}

\begin{document}
\title{Discriminative Local Sparse Representations for Robust Face Recognition}

\author{Yi Chen, Umamahesh Srinivas, Thong T. Do, Vishal Monga, and Trac D. Tran\\
\thanks{Y. Chen, T. T. Do and T. D. Tran are with The Johns Hopkins University, Baltimore, MD, USA. U. Srinivas and V. Monga are with the Pennsylvania State University, University Park, PA, USA. This work has been supported in part by the National Science Foundation (NSF) under Grants CCF-1117545 and CCF-0728893;
the Army Research Office (ARO) under Grant 58110-MA-II and Grant 60219-MA; 
and the Office of Naval Research (ONR) under Grant N102-183-0208.}}


\maketitle%
%
%

\begin{abstract}
A key recent advance in face recognition models a test face image as a \emph{sparse} linear combination of a set of training face images.
The resulting sparse representations have been shown to possess robustness against a variety of distortions like random pixel corruption, occlusion and disguise. This approach however makes the restrictive (in many scenarios) assumption that test faces must be perfectly aligned (or registered) to the training data prior to classification. In this paper, we propose a simple yet robust \emph{local block-based sparsity model}, using adaptively-constructed dictionaries from local features in the training data, to overcome this misalignment problem. Our approach is inspired by human perception: we analyze a series of local discriminative features and combine them to arrive at the final classification decision. We propose a \emph{probabilistic graphical model} framework to explicitly mine the conditional dependencies between these distinct sparse local features. In particular, we learn discriminative graphs on sparse representations obtained from distinct local slices of a face. Conditional correlations between these sparse features are first discovered (in the training phase), and subsequently exploited to bring about significant improvements in recognition rates. Experimental results obtained on benchmark face databases demonstrate the effectiveness of the proposed algorithms in the presence of multiple registration errors (such as translation, rotation, and scaling) as well as under variations of pose and illumination.

\end{abstract}


\section{Introduction}
\label{section::intro}

The problem of automatically verifying the identity of a certain person using a live face capture and comparing against a stored database of human face images has witnessed considerable research activity over the past two decades.
The rich diversity of facial image captures, due to varying illumination conditions, spatial resolution, pose, facial expressions, occlusion and disguise, offers a major challenge to the success of any automatic human face recognition system. A comprehensive survey of face recognition methods in literature is provided in \cite{Zhao_2003}.

In face recognition, indeed any image-based classification problem in general, representative features are first extracted from images typically via projection to a feature space. A classifier is then trained to make class assignment decisions using features obtained from a set of training images. One of the most popular dimensionality-reduction techniques used in computer vision is principal component analysis (PCA). In face recognition, PCA-based approaches have led to the use of eigenpictures\cite{Sirovich_1987} and eigenfaces\cite{Turk_1991} as features. Other approaches have used local facial features \cite{Zou07} like the eyes, nose and mouth, or incorporated geometrical constraints on features through structural matching. An important observation is that different (photographic) versions of the same face approximately lie in a linear subspace of the original image space \cite{Shashua_1992,Belhumeur_1997,Liu01,Basri_2003}. A variety of classifiers have been proposed for face recognition, ranging from template correlation to nearest neighbor and nearest subspace classifiers, neural networks and support vector machines (SVM)~\cite{Zhao_2003}.

Recently, the merits of exploiting parsimony in signal representation and classification have been demonstrated in \cite{Wright_2009,Pillai_2011,Hang_2009}. In their seminal work, Wright {\em et al.} \cite{Wright_2009} argue that a test face image approximately lies in a low-dimensional subspace spanned by (lexicographically ordered) training images themselves. If sufficient training is available, a new test face image has a naturally sparse representation in this overcomplete basis. The sparse vector can be obtained via many norm minimization techniques and is then employed directly for recognition by computing a class (face) specific reconstruction error. Note that there is no {\em offline} training stage in sparsity based face recognition \cite{Wright_2009}, instead the training samples in the dictionary are used directly at the time of testing/recognizing a test face image. The dictionary may be expanded hence as more training (variants of a face image) becomes available. This sparsity-based face recognition algorithm has been shown \cite{Wright_2009} to yield markedly improved recognition performance over traditional efforts in face recognition under various conditions, including illumination, disguise, occlusion, and random pixel corruption.

In many real world scenarios, test images for identification obtained by face detection algorithms are not perfectly registered with the training samples in the databases. The sparse subspace assumption in~\cite{Wright_2009}, however, requires the test face image to be well aligned to the training data prior to classification. Recent approaches have attempted to address this misalignment issue in sparsity-based face recognition\cite{Huang_2008,Wagner_2009,Wagner:robust_face},
usually by jointly optimizing the registration parameters and sparse coefficients and thus leading to more complex systems.

It is well known that, compared to global features, local features may contain more crucial information for representation in many signal and image processing applications. One such example is the block-based motion estimation technique which has been successfully employed in multiple popular video compression standards.

Inspired by the success of locality in recognition, our proposal is the development and use of {\em sparse local features} for face recognition\footnote{Part of this material has been presented in IEEE ICIP 2010\cite{Chen_2009}.}. As our first contribution, we propose a robust yet simpler approach to handle the misalignment problem via a \emph{local block-based sparsity model}.  We are motivated by the observation that a block in the test image can be sparsely represented by a linear combination of blocks in the training images within a spatially-neighboring region, and the sparse representation contains the identity information for the block. The final class decision relies on a combination of decisions from multiple local sparse representations (as observed earlier, the more discriminative facial features such as eyes, nose and mouth constitute a good set of local features). This approach exploits the capability of the local sparsity model to capture relatively stationary features under different types of variations and registration errors.

The presence of multiple feature representations (i.e., the distinct local features) naturally leads to the question: how can we combine the decisions based on multiple local features into a global class decision in the best way possible? A variety of heuristic classifier fusion schemes have been proposed in literature (see \cite{Kittler98} for example). The outputs of individual classifiers constitute \emph{high-level} features. It is reasonable to expect better classification performance by directly exploring the correlation between \emph{low-level} features. In order to explicitly mine such conditional dependencies between these distinct sparse local features, we propose a probabilistic graphical model framework as the second main contribution of this paper\footnote{Part of this material has been accepted to IEEE Asilomar Conf. 2011\cite{faceRec:Srinivas11}.}. In particular, we learn discriminative graphs on sparse representations obtained from distinct local slices of a face. Conditional correlations between these sparse features are first discovered by learning discriminative tree graphs \cite{Tan10} on each distinct feature set. The initial disjoint trees are then thickened, i.e., augmented with more edges to capture newly learnt feature correlations, via boosting \cite{freund:Adaboost} on disjoint graphs. Probabilistic graphical models offer additional benefits in terms of robustness to limited training, and reduced computational complexity of inference.

It is informative to contrast our contribution with recent work in robust face recognition that considers registration errors. Huang \emph{et al.}\cite{Huang_2008} consider the scenario where the test images can be represented in terms of all training images and (linearized versions of) their image plane transformations. The computational cost scales with the complexity of the plane transformation. In \cite{Wagner:robust_face}, the difficult nonconvex problem of simultaneous optimization over sparse coefficients and registration parameters is relaxed via sequential iterative minimization. In addition, a novel projector-based illumination system has been proposed to capture variations in scene lighting. In our proposed approach however, the registration parameters are not explicitly determined. Instead, robustness to misalignment is introduced by augmenting the training with spatially local blocks from each training image. Another significant departure from existing sparse representation-based approaches is our use of a principled strategy via graphical models to explicitly mine feature dependencies, instead of performing classification using only reconstruction residuals.

The rest of this paper is organized as follows. Section \ref{section::background} provides a review of  sparsity based face recognition, as well as an overview of probabilistic graphical models. The two main contributions of this paper are presented in Section~\ref{section::method}. An extensive set of experiments is performed on popular face recognition databases to validate the effectiveness of our proposed framework, and results under varying practical settings are provided in Section~\ref{section::simulation}.  Section~\ref{section::conclusion} summarizes the contributions and concludes the paper.

\section{Background}
\label{section::background}

\subsection{Sparse Representation-based Classification}
\label{section::src_review}
As mentioned earlier, algorithmic advances in face recognition have been comprehensively surveyed in the literature \cite{Zhao_2003}. Here, we primarily review recent pioneering work in sparse representation-based face recognition \cite{Wright_2009}, which forms the foundation for our proposed contribution. This method advocates the use of sparse representation in a \emph{discriminative} setting, a novel advance over previous work which exploited sparsity from a \emph{signal recovery} standpoint.

First, let us introduce the standard notation that will be used throughout this paper. Suppose there are $K$ different classes (corresponding to unique faces), labeled ${C}_1,\ldots,{C}_K$, and there are $N_i$ training samples (each in $\R^n$) corresponding to class ${C}_i, i = 1,\ldots,K$. The training samples corresponding to class ${C}_i$ can be collected in a matrix $\mat D_i \in \R^{n\times N_i}$, and the collection of all training samples is expressed using the matrix:
\be
\mat D = [\mat D_1 ~ \mat D_2 ~\ldots ~ \mat D_K],
\label{eq::basis_rep}
\ee
where $\mat D \in \R^{n \times T}$, with $T = \sum_{k=1}^{K}N_k$. A new test sample $\mb{y} \in \R^{n}$ can be expressed as a sparse linear combination of the training samples:
\be
\vect y = \mat D \bm{\alpha},
\label{eq::sparse_rep}
\ee
where $\bm{\alpha}$ is expected to be a sparse vector (i.e., only a few entries in $\bm{\alpha}$ are nonzero). This is an underdetermined system of linear equations. The classifier seeks the sparsest representation by solving:
\be
\hat{\bm{\alpha}} = \arg\min \norm{\bm{\alpha}}_0 \quad\text{subject to}\quad \|\mat D\bm{\alpha} -\mb{y}\|_2 \leq \epsilon,
\label{eqn::global_l0}
\ee
where $\norm{\cdot}_0$ denotes the number of nonzero entries in the vector. Under a set of sufficient conditions (that hold in general for the above problem set-up), it has been shown theoretically \cite{CandesRob06} that the non-convex optimization problem represented by \eqref{eqn::global_l0} can be relaxed to the following convex optimization problem:
\be
\hat{\bm{\alpha}} = \arg\min \norm{\bm{\alpha}}_1 \quad\text{subject to}\quad \|\mat D\bm{\alpha} -\mb{y}\|_2 \leq \epsilon.
\label{eqn::global_l1}
\ee
Alternatively, the problem in~(\ref{eqn::global_l0}) can be solved by greedy pursuit algorithms~\cite{Tropp_2005,Dai_2009, Do_2008}.

Once the sparse vector is recovered, the identity of $\mb{y}$ is given by the minimal residual
\be
\label{eqn::global_identity}
\text{identity}(\mb{y}) = \arg\min_i\norm{\mb{y}  - \mat D \delta_i(\hat{\bm{\alpha}})},
\ee
where $\delta_i(\bm{\alpha})$ is a vector whose only nonzero entries are the same as those in $\bm{\alpha}$ but only associated with
class ${C}_i$. The particular choice of class-specific residuals makes the task of decision assignment computationally trivial.

Often, it is necessary to check if a particular test image belongs to any of the available classes. The authors develop a sparsity concentration index (SCI) to decide if a test image is valid or not. Given a sparse coefficient vector $\bm{\alpha} \in \R^{T}$, the SCI is defined as follows:
\be
\mbox{SCI}(\bm{\alpha}) = \frac{K \cdot \max_i \|\delta_i(\bm{\alpha})\|_1/\|\bm{\alpha}\|_1 - 1}{K - 1} \in [0,1].
\label{eq::sci}
\ee
A high value of SCI indicates a sparse representation corresponding to a valid test image, while a value close to 0 indicates that the feature coefficients are distributed across all classes.

\subsection{Probabilistic Graphical Models}
\label{section::gm_review}

We provide a brief overview of probabilistic graphical models from an inference (hypothesis testing) viewpoint. Discriminative graphs will be used to model the class conditional densities $f(\bm \alpha | C_{i})$, i.e., a set of p.d.fs defined on the (sparse) coefficient vector which are employed to make class assignments (each class $C_i$ corresponds to the $i$-th person in the database).

A graph $\mathcal{G} = (\mathcal{V},\mathcal{E})$ is defined by a collection of nodes $\mathcal{V} = \{v_1,\ldots,v_r\}$ and a set of (undirected) edges $\mathcal{E} \subset \binom{\mathcal{V}}{2}$, i.e., the set of unordered pairs of nodes. A probabilistic graphical model is obtained by defining a random vector on $\mathcal{G}$ such that each node represents one (or more) random variables and the presence of edges indicates conditional dependencies. The graph structure thus enforces a  particular factorization of the joint probability distribution in terms of pairwise marginals.

The use of graphical models in applications has been motivated by practical concerns like insufficient training to learn models for high-dimensional data and the need for reduced computational complexity in realtime tasks \cite{Lauritzen96,wainwright:GMs}. Graphical models offer an alternate visualization of a probability distribution from which conditional dependence relations can be easily identified. Graphical models also enable us to draw upon the rich resource of efficient graph-theoretic algorithms to learn complex models and perform inference.

Graphical models can be learnt from data in two different settings: generative and discriminative. In generative learning, a \emph{single} graph is learnt to approximate a given distribution by minimizing a measure of \emph{approximation error}. Generative learning approaches trace their origin to Chow and Liu's \cite{Chow68} idea of learning the optimal tree approximation $\hat{p}$ of a multivariate distribution $p$ using first- and second-order statistics:
\be
\hat{p} = \arg \min_{\hat{p}~\mbox{\scriptsize{is a tree}}} D(p||\hat{p}),
\label{eq:chow-liu}
\ee
where $D(p||\hat{p}) = E_{p}[\log (p/\hat{p})]$ denotes the Kullback-Leibler (KL) divergence. This optimization problem is
shown to be equivalent to a maximum-weight spanning tree (MWST) problem. In discriminative learning, on the other hand, a pair of graphs is jointly learnt from a pair of empirical estimates by minimizing the \emph{classification error}. (Note that we consider binary classification problems here to reduce notational clutter. The approach naturally extends to multi-class problems by learning graphs in a one-versus-all manner.)

Recently, Tan {\em et al.}\cite{Tan10} proposed a graph-based discriminative learning framework, based on maximizing an approximation to the $J$-divergence, which is a symmetric extension of the KL-divergence. Given two probability distributions $p$ and $q$, their $J$-divergence is defined as: $J(p,q) = D(p||q) + D(q||p)$. The tree-approximate $J$-divergence is then defined as:
\be
\hat{J}(\hat{p},\hat{q};p,q) = \int (p(x)-q(x))\log\left[\frac{\hat{p}(x)}{\hat{q}(x)}\right] dx,
\label{eq:J-approx}
\ee
which measures the ``separation'' between tree-structured approximations $\hat{p}$ and $\hat{q}$. Using the key observation that maximizing the $J$-divergence minimizes the upper bound on probability of classification error, the discriminative tree learning problem is then stated (in terms of empirical estimates $\tilde{p}$ and $\tilde{q}$) as follows:
\be
(\hat{p},\hat{q}) = \arg \max_{\hat{p},\hat{q}~\mbox{\scriptsize{trees}}} \hat{J}(\hat{p},\hat{q};\tilde{p},\tilde{q}).
\label{eq:disc-learn}
\ee
It is shown in \cite{Tan10} that this optimization further decouples into two MWST problems:
\bea
\hat{p} & = & \arg \min_{\hat{p}~\mbox{\scriptsize{tree}}} D(\tilde{p}||\hat{p}) - D(\tilde{q}||\hat{p}) \label{eq:p-opt}\\
\hat{q} & = & \arg \min_{\hat{q}~\mbox{\scriptsize{tree}}} D(\tilde{q}||\hat{q}) - D(\tilde{p}||\hat{q}) \label{eq:q-opt}.
\eea

Here, \eqref{eq:p-opt} and \eqref{eq:q-opt} bring out the distinction (from a classification viewpoint) between: (i) using generative learning techniques to separately learn $\hat{p}$ and $\hat{q}$ and then performing inference, and (ii) simultaneously learning a pair of graphs discriminatively. In \eqref{eq:p-opt}, the optimal $\hat{p}$ is chosen to minimize its (KL-divergence) distance from $\tilde{p}$ and \emph{simultaneously} maximize its distance from $\tilde{q}$.

The discussion so far mainly considers tree graphs, which are fully connected acyclic graphical structures. The computational burden of learning tree graphs is significantly reduced owing to the sparse connectivity. This feature however imposes a limitation on the complexity of the model so learnt. This inherent trade-off between generalization and performance poses a serious challenge to the application of graphical models in various tasks.

Our contribution as described in the remainder of this paper proposes an extension of discriminative graph learning for the purpose of face recognition, utilizing distinct local features from a block-based sparsity model.

\section{Face Recognition Via Local Decisions From Locally Adaptive Sparse Features}
\label{section::method}


The two main contributions of this paper are presented in Sections \ref{section::contrib1} and \ref{section::contrib2} respectively. Section \ref{section::contrib1} explains the process of obtaining local sparse features. In Section \ref{section::contrib2}, two different methods of combining class decisions are proposed: (i) based on reconstruction error, and (ii) using graphical models.

\begin{figure}[htp]
\centering
\includegraphics[width=.8\columnwidth]{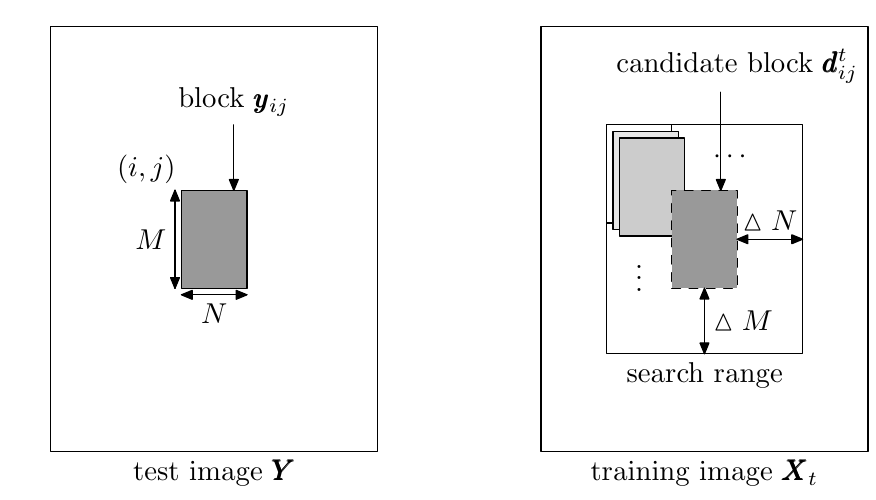}\\(a)\\
\vspace{3pt}
\includegraphics[width=.8\columnwidth]{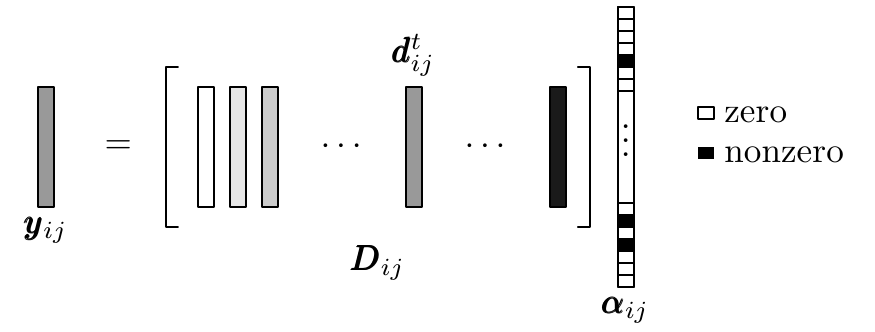}\\(b)
\caption{Representation of a block in the test image from a locally
adaptive dictionary. (a)~The blocks in the test and training images
(only one training sample is displayed). (b)~Sparse representation
$\vect y_{ij} = \mat D_{ij} \vect\alpha_{ij}$.}
\label{fig::model}
\end{figure}

\subsection{Locally Adaptive Sparse Representations}
\label{section::contrib1}

The method in \cite{Wright_2009} advances practical face recognition by enabling significantly enhanced robustness to distortions like occlusion, pixel corruption and disguise. However, as discussed in Section \ref{section::intro}, the subspace model requires precise registration making their approach vulnerable to alignment errors of rotation, translation and scaling that are natural to face capture processes. To deal specifically with disguise, Wright {\em et al.} \cite{Wright_2009} do suggest a block-partitioning scheme which to a first order captures local face image characteristics while still suffering from misalignment. The superior compression ability of local features compared to global representations is also well-known from applications like block-based motion estimation in video coding. In other words, local sparsity is beneficial from the recovery standpoint. In this work, we consummate this intuition by designing adaptive dictionaries for each ``local block" such that the resulting (local) sparse representations naturally exhibit robustness to alignment errors.

To achieve this, in the proposed local sparse representation model for face recognition, we adopt the inter-frame sparsity model in~\cite{Do_2009}, where a block in a video frame is expressed as a sparse linear combination of
a few spatially-adjacent blocks from the reference frames.
An illustration of the proposed model in shown in Fig.~\ref{fig::model}, where a block in the (possibly misaligned) test image $\mat Y$ is sparsely represented by a locally adaptive dictionary consisting of blocks in the training images $\left\{\mat X_t\right\}_{t=1,\ldots,T}$ within the same spatial neighborhood.
Note that for illustration, only one training image is shown in~Fig.~\ref{fig::model}(a).
Specifically, let $\vect y_{ij}\in\R^{MN}$ be the vectorized $M\times N$ block in the test image $\mat Y$ with the upper left pixel located at $(i,j)$.
The search region $\mat S^t_{ij}$ in the $t$-th training image $\mat X_t$ is an $(M+2\vartriangle M)\times(N+2\vartriangle N)$ image region:
$$\mat S^t_{ij} = \begin{bmatrix}
  x^t_{i-\vartriangle M,j-\vartriangle N} & \cdots & x^t_{i-\vartriangle M,j+N-1+\vartriangle N}\\
  \vdots & \ddots & \vdots\\
  x^t_{i+M-1+\vartriangle M,j-\vartriangle N} & \cdots & x^t_{i+M-1+\vartriangle M,j+N-1+\vartriangle N}\\
\end{bmatrix}.$$
The local dictionary $\mat D_{ij}$ for the block $\vect y_{ij}$ is then constructed by all $M\times N$ blocks within the search regions $\left\{\mat S^t_{ij}\right\}_{t = 1,2, \ldots, T}$ in the $T$ training images:
$$\mat D_{ij} = \begin{bmatrix}
    \mat D^1_{ij} & \mat D^2_{ij} & \cdots & \mat D^T_{ij}
\end{bmatrix},$$
where each $$\mat D^t_{ij} = \begin{bmatrix}
  \vect d^t_{i-\vartriangle M,j-\vartriangle N}&\vect d^t_{i-\vartriangle M,j-\vartriangle N+1}&\cdots&\vect d^t_{i+\vartriangle M,j+\vartriangle N}
\end{bmatrix}$$ is an $(MN)\times\big((2\vartriangle M+1)(2\vartriangle N+1)\big)$ sub-dictionary whose columns are the vectorized blocks in the $t$-th training image defined in the same way as $\vect y_{ij}$.

In this way, a locally-adaptive dictionary $\mat D_{ij}$ is constructed for every block of interest in the test image.
The size of the dictionary depends on the non-stationary behavior of the data as well as the level of computational complexity we can afford.
For significant registration errors, the local dictionaries can be augmented by distorted versions of the local blocks in the training data for better performance at the cost of higher computational intensity.
Compared to the original global approach, the dictionary $\mat D_{ij}$ captures local characteristics better and yields a reasonable approximation of the training image at the block level. Our approach is different from patch-based dictionary learning~\cite{Elad_2006} in multiple aspects: (i) we emphasize the local adaptivity of the dictionaries, and (ii) our dictionaries are constructed by simply taking blocks from training data without any sophisticated learning process.

\begin{figure}[thp]
\centering
\begin{tabular}{cc}
\includegraphics[scale=0.5]{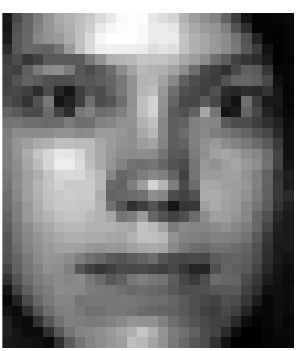} & \includegraphics[scale=0.5]{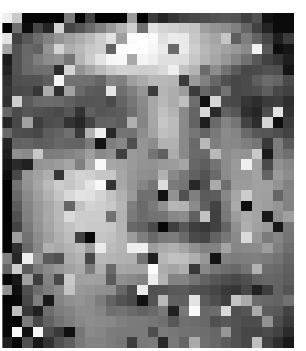}\\
(a) & (b)
\end{tabular}
\includegraphics[width=\columnwidth,height=.15\columnwidth]{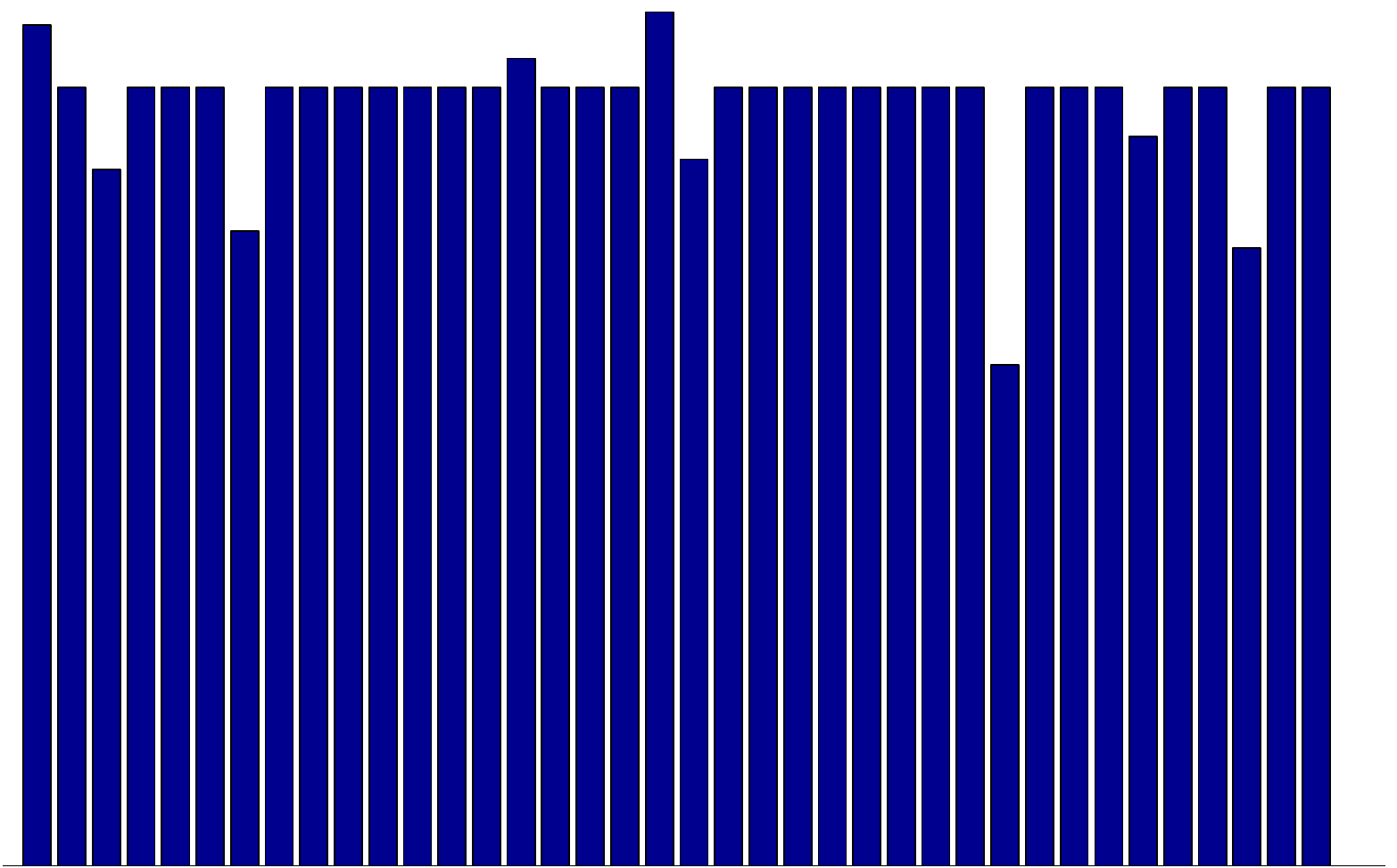}\\
(c)
\includegraphics[width=\columnwidth]{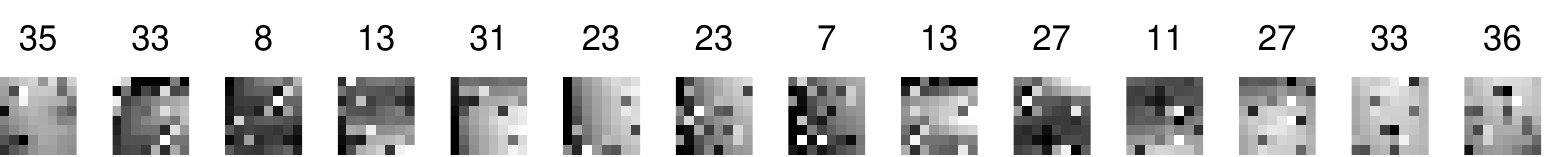}\\
\includegraphics[width=\columnwidth]{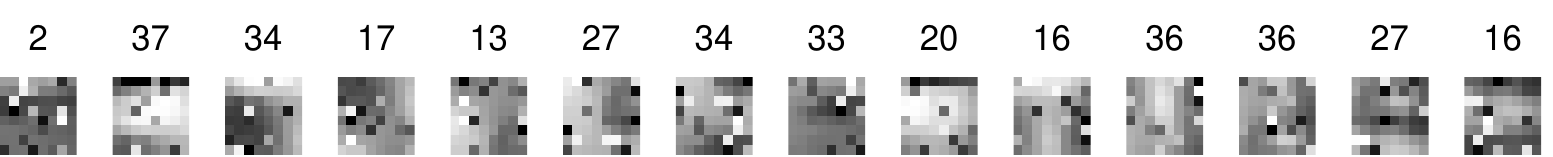}\\
\includegraphics[width=\columnwidth]{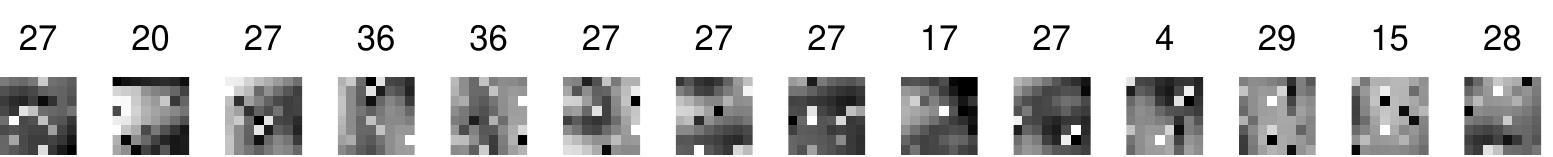}\\
(d)\\
\includegraphics[width=\columnwidth,height=.15\columnwidth]{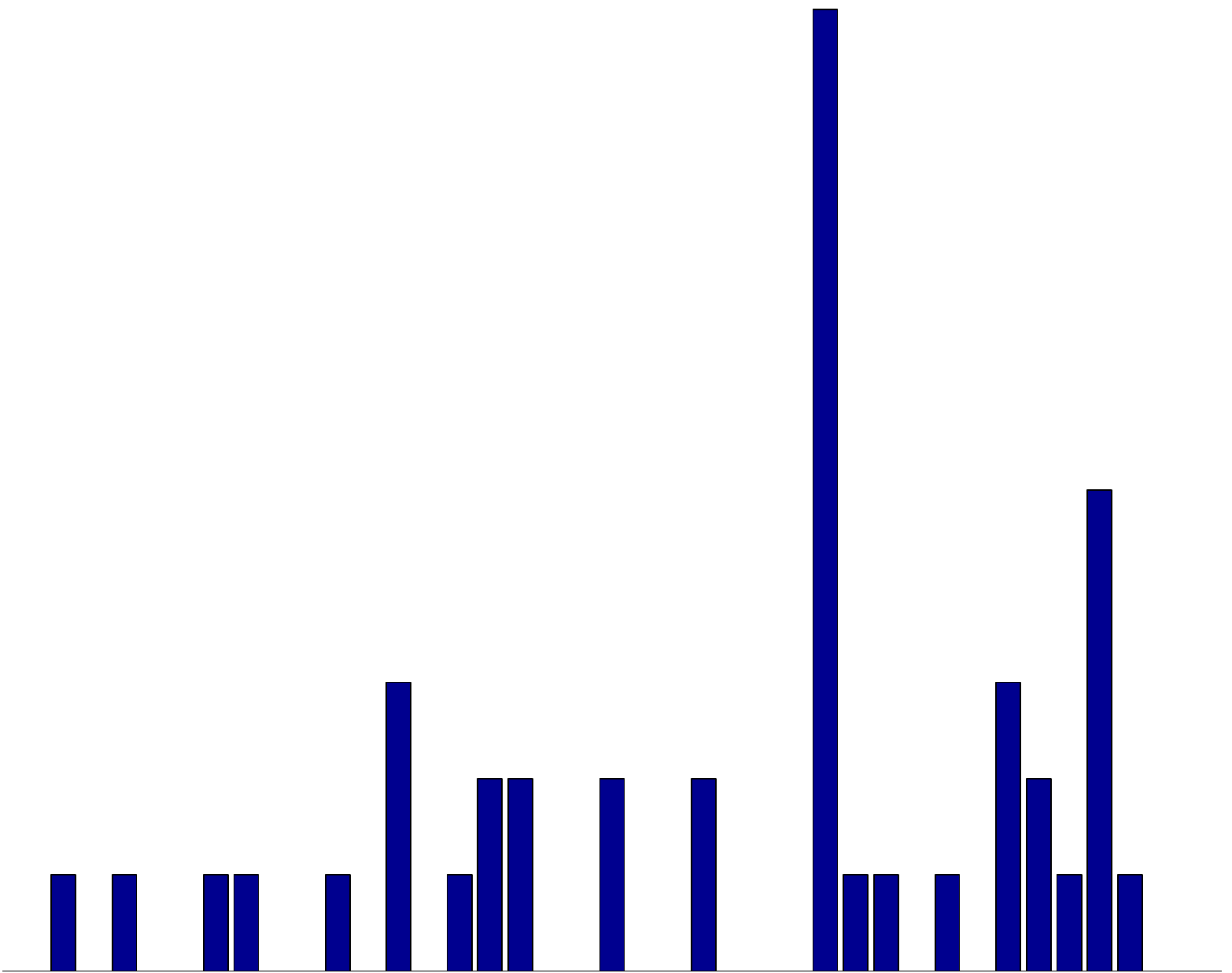}\\
(e)\\
\includegraphics[width=\columnwidth,height=.15\columnwidth]{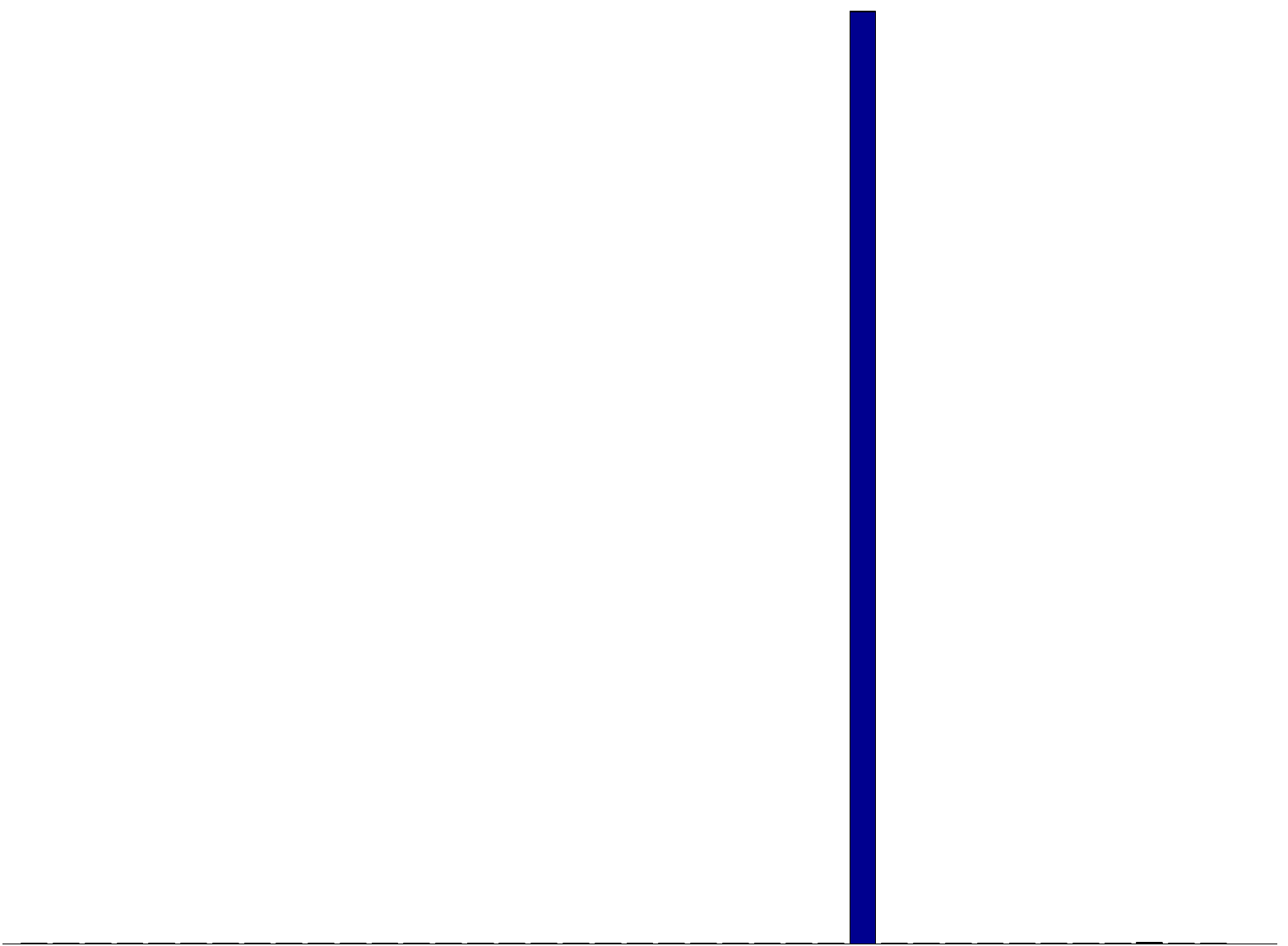}\\
(f) \caption{Example of the proposed sparsity-based approach using
multiple test blocks. (a)~Original image (Class 27). (b)~Distorted test image $\mat
Y$. (c)~Residuals using the original global approach:
identity$(\vect Y) = 29$. (d)~Classification results for each of the
42 blocks $\left\{\vect y_l\right\}_{l=1,\ldots,42}$. (e)~Number of
votes for the $k$th class, $k = 1, \ldots, 38$. Identity$(\vect Y) =
27$. (f)~Probability of $\left(\text{identity}(\mat Y) = k\right)$,
$k = 1, \ldots, 38$. Identity$(\vect Y) = 27$.}
\label{fig::example}
\end{figure}

We propose that the block $\vect y_{ij}$ in the misaligned image $\mat Y$ can be approximated by a linear combination of only a few atoms in the dictionary $\mat D_{ij}$:
\be
\label{eqn::local_sparsity}
\vect y_{ij} = \mat D_{ij} \vect \alpha_{ij},
\ee
where $\vect \alpha_{ij}$ is a sparse vector, as illustrated in Fig.~\ref{fig::model}(b).
Similar to the global case, the sparse vector is recovered by solving the following optimization problem:
\be
\label{eqn::local_l0}
\hat{\vect \alpha}_{ij} = \arg\min \norm{\vect \alpha_{ij}}_0 ~\text{subject to}~ \|\mat D_{ij} \vect \alpha_{ij} - \vect y_{ij}\|_2 \leq \epsilon.
\ee
Note that the resulting complexity of the overall algorithm is still manageable since the above sparse recovery is performed on a small block with a dictionary of modest size.
After the sparse vector $\hat{\vect \alpha}_{ij}$ is obtained, the error residual with respect to the $k$-th class sub-dictionary is calculated by
\be
\label{eqn::local_residual}
r^k(\vect y_{ij}) = \norm{\vect y_{ij}-\mat D_{ij} \vect \delta_k\left(\hat{\vect \alpha}_{ij}\right)}_2,
\ee
where $\vect \delta_k\left(\hat{\vect \alpha}_{ij}\right)$ is as defined in~(\ref{eqn::global_identity}).
Then, the identity of the test block can be determined by the minimal residual as follows:
\be
\label{eqn::local_identity}
\text{identity}(\vect y_{ij}) = \arg\min_{k = 1,\ldots,K} r^k(\vect y_{ij}).
\ee

The usage of a single block certainly cannot produce outstanding classification performance. To improve the algorithm's robustness, we employ multiple blocks: solving the sparse recovery problem for each block individually, and then combining the results for all of the blocks. Blocks may be chosen manually in the areas with discriminative features (such as eyes, nose, and mouth), or areas with high SNR/more variations, or uniformly in the entire test image in non-overlapped or overlapped fashion. It should be noted that the blocks can be processed independently in parallel. Moreover, since blocks can be overlapped, our approach is computationally scalable - more computation simply delivers better recognition performance - a feature that will be illustrated by experimental results in Section~\ref{section::simulation}.

Finally, we would like to remark that our locally adaptive sparse representation is a more general and more powerful framework comparing to the global sparse representation proposed in~\cite{Wright_2009}.
In other words, if we set the parameters $\vartriangle M$ and $\vartriangle N$ to zero, and further force the local sparse vectors $\vect \alpha_{ij}$ to be the same for all non-overlapped test block $\vect y_{ij}$, then what we get back is essentially the global sparse representation.

\subsection{Recognition Decisions from Local Sparse Features}
\label{section::contrib2}
\subsubsection{Classifiers based on reconstruction error}
We first present two simple schemes that combine the individual recognition results from the blocks. Let $\left\{\vect y_l\right\}_{l=1,\ldots,L}$ be the $L$ blocks in the test image $\mat Y$. (Note that in Section \ref{section::contrib1}, we have identified each block with the location $(i,j)$ of its upper left pixel. Here, each block identifier $l$ is implied to have unique correspondence with one such pixel location, and we will use $\vect y_l$ instead of $\vect y_{ij}$ henceforth.)

\paragraph{Majority voting}
\be
\label{eqn::voting}
\text{identity}\left(\mat Y\right) = \max_{k = 1,\ldots,K}\left|\left\{l = 1,\ldots,L: \text{identity}\left(\vect y_l\right)=k\right\}\right|,
\ee
where $|S|$ denotes the cardinality of a set $S$ and $\text{identity}\left(\vect y_l\right)$ is determined by (\ref{eqn::local_identity}).

\paragraph{Maximum likelihood} This is another intuitive approach of fusing classification results from multiple blocks. Let $\hat{\vect \alpha}_l$ be the recovered sparse representation vector of the block $\vect y_l$ and the local dictionary $\mat D_l$. We define the probability of $\vect y_l$ belonging to the $k$-th class to be inversely proportional to the residual associated with the dictionary atoms in the $k$-th class:
\be
\label{eqn::prob}
p^k_l = P\left(\text{identity}\left(\vect y_l\right)=k\right) = \frac{1/r^k_l}{\sum_{k=1}^K\left(1/r^k_l\right)},
\ee
where $r^k_l = \norm{\vect y_{l}-\mat D_{l} \vect \delta_k\left(\hat{\vect \alpha}_{l}\right)}_2$ is the residual associated with the $k$-th class as in \eqref{eqn::local_residual}. The identity of the test image $\mat Y$ is then given by
\be
\label{eqn::local_prob_combine}
\text{identity}\left(\mat Y\right) = \arg\max_{k=1,\ldots,K}\log\left(\prod_{l=1}^L p^k_l\right).
\ee
The likelihood measure can also be used as a criterion for outlier rejection, since the probability of an outlier belonging to a particular class tends to be uniformly distributed among all training classes.

An example of the proposed approach fusing results of multiple local blocks is illustrated in Fig.~\ref{fig::example} using the Extended Yale B Database~\cite{GeBeKr01}, which consists of facial images of 38 individuals. More details about experiments will be discussed in Section \ref{section::simulation}. Fig.~\ref{fig::example}(a) shows an image belonging to the 27th class, and Fig.~\ref{fig::example}(b) shows the test image to be classified, which is the image in~(a) distorted by rotation, scaling, and random pixel corruption. The distortion causes the failure of the original global approach in~\cite{Wright_2009} in this case, as seen by the error residuals in Fig.~\ref{fig::example}(c) where the 29th class turns out to yield the minimal residual. For the proposed local approach, we use 42 blocks of size $8\times 8$ chosen uniformly from the distorted test image. The blocks and class labels for each individual block are displayed in Fig.~\ref{fig::example}(d). Figs.~\ref{fig::example}(e) and~(f) show the number of votes and the probability defined in \eqref{eqn::prob}, respectively. It is obvious that in both cases, the local approach yields the correct class label (i.e., the 27th class has the highest number of votes and the maximal probability). This example also highlights the robustness of local sparse representations under reduced feature dimensions, although the individual blocks are chosen uniformly instead of selectively corresponding to representative facial features.

\begin{figure}[t]
\begin{center}
\includegraphics[scale=0.27]{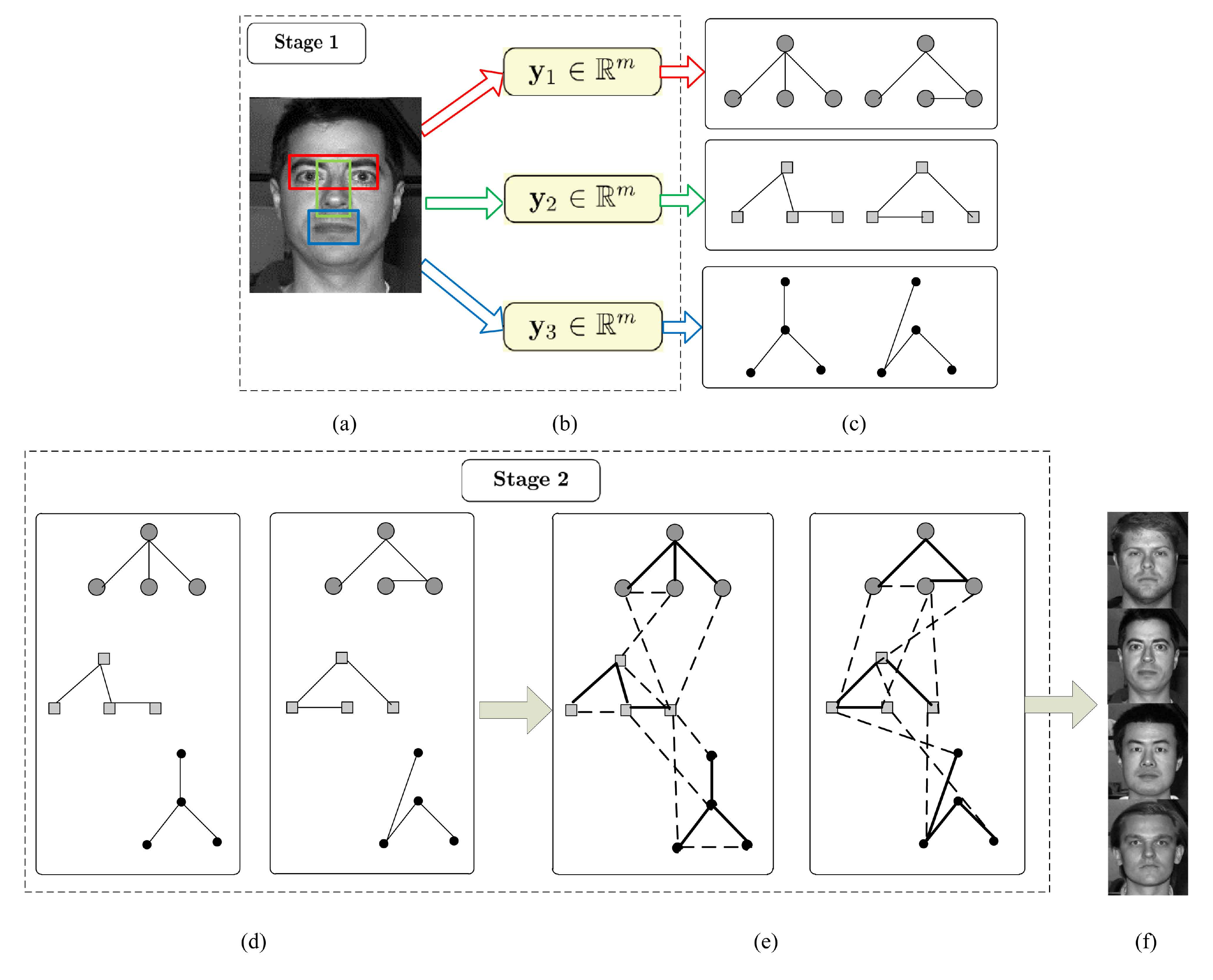}
\vspace{-5mm}
\caption{Proposed framework for face recognition: (a) Target face image, (b) Local regions for extracting sparse features, (c) Initial pairs of \emph{tree} graphs for each feature set, (d) Initial sparse graph formed by tree concatenation, (e) Final pair of thickened graphs; newly learned edges represented by dashed lines, (f) Graph-based inference. In (c)-(e), the graphs on the left and right correspond to distributions $p$ (class $C_i$) and $q$ (class $\tilde{C}_i$) respectively.}
\label{fig:igt}
\end{center}
\vspace{-6mm}
\end{figure}

\subsubsection{Graphical models to mine feature correlations}
\label{section:gm_class}

The two schemes discussed above, albeit intuitively motivated, are essentially heuristic ways of fusing classifier outputs. We now present a two-stage probabilistic graphical model framework to directly exploit conditional correlations between features from local regions themselves. The overall framework is shown in Fig. \ref{fig:igt}.

We introduce some additional notation. Let ${C}_i, i = 1,2,\ldots,K$ denote the $i$-th class of face images (as defined earlier), and let $\tilde{{C}_i}$ denote the class of face images complementary to class ${C}_i$, i.e., $\tilde{{C}_i} = \bigcup_{k=1,\ldots,K, k \neq i} {C}_k$. Let $\mathcal{B}_i$ denote the $i$-th binary classification problem of classifying a query face image (or corresponding feature) into ${C}_i$ or $\tilde{{C}_i}$ ($i = 1,\ldots,K$). As will be clear shortly, defining $K$ such binary problems is necessary for application of the discriminative graphical framework. The graphical model-based algorithm is summarized in Algorithm \ref{alg:igt}, and it consists of an {\em offline} stage to learn the discriminative graphs (Steps 1-4) followed by an {\em online} stage (Steps 5-6) where a new test image is classified.

The offline stage involves extraction of features from training images, which comprise the empirical estimates from which approximate p.d.fs for each class are learnt after the graph thickening procedure. The individual steps in this stage are explained next.
\paragraph{Feature extraction} Let us first consider one of the local spatial regions in the face, say corresponding to the eyes. For the binary classification problem $\mathcal{B}_i$, dictionaries $\mat D_i$ and $\tilde{\mat D}_i$ are constructed according to the procedure in Section \ref{section::contrib1}, using samples from $C_i$ and $\tilde{C}_i$ respectively. (Subscripts are dropped while denoting the dictionaries to avoid confusion, and they can be inferred from context.) Features in $\mathbb{R}^m$ are now extracted for any block $\vect z$ (spatially corresponding to eyes) by solving the sparse recovery problem:
\be
\hat{\vect \beta} = \arg\min \norm{\vect \beta}_0 ~\text{subject to}~ \|\mat D \vect \beta - \vect z\|_2 \leq \epsilon,
\label{eqn::sparse_ftr_ext}
\ee
where $\mat D:= [\mat D_i, ~ \tilde{\mat D}_i]$. Features corresponding to other local regions are generated analogously. Training features (that form the overcomplete dictionary) for ${C}_i$ are obtained by using training faces that are {\em known} to belong to class ${C}_i$, while features for $\tilde{{C}}_i$ are obtained by choosing representative training from $\tilde{{C}}_i$ as input to the feature extraction process.

\begin{algorithm}[t]
\caption{Discriminative graphical models for face recognition (Steps 1-4 offline)}
\label{alg:igt}
\begin{algorithmic}[1]
\STATE \textbf{Feature extraction (training):} Obtain sparse representations $\bm{\alpha}_l,l=1,\ldots,P$ in ${\mathbb R}^{m}$ from facial features, using adaptive locally block-sparsity model \eqref{eqn::sparse_ftr_ext}
\STATE \textbf{Initial disjoint graphs:}\\
       For $l = 1,\ldots,P$\\
       Discriminatively learn pairs of $m$-node tree graphs $\mathcal{G}_l^p$ and $\mathcal{G}_l^q$ on $\{\bm{\alpha}_l\}$ obtained from training data
\STATE Separately concatenate nodes corresponding to $p$ and $q$ respectively, to generate initial graphs
\STATE \textbf{Boosting on disjoint graphs}: Iteratively thicken initial disjoint graphs via boosting to obtain final graphs $\mathcal{G}^p$ and $\mathcal{G}^q$
\hrule
\COMMENT{\textbf{Online process}}
\STATE \textbf{Feature extraction (test):} Obtain sparse representations $\bm{\alpha}_l,l=1,\ldots,P$ in ${\mathbb R}^{m}$ from test image
\STATE \textbf{Inference}: Classify based on output of the resulting classifier using \eqref{eq:decision_rule}.
\end{algorithmic}
\end{algorithm}

\paragraph{Initial disjoint pairs of trees}
The extraction of different sets of features from input face images is performed offline. Each such representation may be viewed as a projection $\mathcal{P}_l: \mathbb{R}^n \mapsto \mathbb{R}^{m}$. In our framework we consider, in all generality,  $P$ distinct projections $\mathcal{P}_l, l = 1,2,\ldots,P$. For every input image $\vect y \in \mathbb{R}^n$, $P$ different features $\bm{\alpha}_l \in \mathbb{R}^{m}, l = 1,2,\ldots, P$ are obtained. Fig.\ \ref{fig:igt}(b) depicts this process for the particular case $P = 3$, i.e., using eyes, nose and mouth as features. The different projections lead to local features that have complementary yet correlated information, since they arise from the same original face image.

Figs.\ \ref{fig:igt}(c)-(f) represent the graph learning process. We denote the class distributions corresponding to $C_i$ and $\tilde{C}_i$ by $p$ and $q$ respectively, i.e., \ $f_{p}^{i}(\bm{\alpha}_l) = f(\bm{\alpha}_l|C_i)$ and $f_{q}^{i}(\bm{\alpha}_l) = f(\bm{\alpha}_l|\tilde{C}_i)$. A pair of $m$-node discriminative tree graphs $\mathcal{G}_l^p$ and $\mathcal{G}_l^q$ is learnt for each projection $\mathcal{P}_l, l = 1,2,\ldots,P$, by solving \eqref{eq:p-opt} and \eqref{eq:q-opt}. The local sparse features $\bm{\alpha}_l$ obtained from the $P$ local blocks are used as empirical estimates to train the tree graphs\footnote{The same training faces present in the overcomplete dictionary are used to generate the sparse features to train the graphs.}. By concatenating the nodes of the graphs $\mathcal{G}_l^p, l = 1,\ldots,P,$ we have one initial sparse graph structure with $Pm$ nodes (Fig. \ref{fig:igt}(d)). Similarly, we obtain another initial graph by concatenating the nodes of the graphs $\mathcal{G}_l^q, l = 1,\ldots,P$. We have now learnt (graphical) p.d.fs $\hat{f}_p^{i}(\bm{\alpha}_l)$ and $\hat{f}_{q}^{i}(\bm{\alpha}_l)$, where $\bm{\alpha}_l$ is the sparse feature vector obtained from the $l$-th local block ($l = 1,\ldots,P$), and $i$ refers to the $i$-th binary classification problem $\mathcal{B}_i$. Inference based on these disjoint graphs can be interpreted as feature fusion assuming statistical independence of the individual target image representations.

\paragraph{Discriminative graphs for classification}
Although simple tree graphs can be learnt efficiently, their ability to model general distributions is limited. However, learning graphs with arbitrarily complex structure is known to be an NP-hard problem \cite{Friedman97}. To overcome this trade-off, we learn different pairs of discriminative graphs over the same sets of nodes (but weighted differently) in different iterations via boosting and obtain a ``thicker'' graph by augmenting the original trees with the newly-learned edges \cite{discGM:Srinivas11}. Boosting \cite{freund:Adaboost} iteratively improves the performance of weak learners into a classification algorithm with arbitrarily accurate performance.

For each binary classification problem, the $P$ pairs of tree graphs in Fig. \ref{fig:igt}(c) are discriminatively learnt \cite{Tan10} from distinct local regions of the face image using empirical estimates of distributions available from corresponding training samples of locally sparse features. In Fig. \ref{fig:igt}(c), an example instantiation is shown for $P = 3$ where the local regions correspond to eyes, nose and mouth respectively.
They are subsequently thickened by the process of boosting \cite{freund:Adaboost}, \cite{discGM:Srinivas11}. This process of learning new edges is tantamount to discovering new conditional correlations between distinct sets of local features, as illustrated by the dashed edges in Fig. \ref{fig:igt}(e). The thickened graphs $\hat{f}_p^{i}(\bm{\alpha})$ and $\hat{f}_q^{i}(\bm{\alpha})$ are therefore estimates of the true (but unknown) class conditional p.d.fs $f_{p}^{i}(\bm{\alpha}) = f(\bm{\alpha}|C_i)$ and $f_{q}^{i}(\bm{\alpha}) = f(\bm{\alpha}|\tilde{C}_i)$, where $\bm{\alpha}$ is the concatenated feature vector from all $P$ blocks.

The graph learning procedure described so far is performed offline. The actual classification of a new test image is performed in an online process, explained next.
\paragraph{Feature extraction} The feature extraction is identical to the process described in the offline stage. Corresponding to each test image, local features $\bm{\alpha}_l, l = 1,\ldots,P$ are obtained by solving the individual sparse recovery problems.

\paragraph{Inference} Classification is performed in a \emph{one-versus-all} manner by solving $K$ separate binary classification problems $\mathcal{B}_i$. If $\hat{f}_p^{i}$ and $\hat{f}_q^{i}$ denote the final probabilistic graphical models learnt for $C_i$ and $\tilde{{C}}_i$ ($i = 1,2,...K$) respectively, then the face image feature vector comprising of sparse coefficients from all the local blocks, i.e., $\bm {\alpha}$ is assigned to a class $i^{*}$ according to the following decision rule:
\be
i^{*} = \arg\max_{i \in \{1,\ldots,K\}}\log\left(\frac{\hat{f}_{p}^{i}(\bm {\alpha})}{\hat{f}_{q}^{i}(\bm {\alpha})}\right).
\label{eq:decision_rule}
\ee

\section{Experiments and Results}
\label{section::simulation}

\begin{table}[t]
\begin{center}
\caption{Overall recognition rates using calibrated test images from the Extended Yale B database (Section \ref{section::calibrated}).}
\begin{tabular}{|c|c|c|c|c|}
  \hline
  Method & Recognition rate ($\%$)\\
  \hline
  LSGM & 97.3 \\
  SRC & 97.1 \\
  Eigen-NS & 89.5 \\
  Eigen-SVM & 91.9\\
  Fisher-NS & 84.7\\
  Fisher-SVM & 92.6 \\
  \hline
\end{tabular}
\label{table::calibrated}
\end{center}
\end{table}

\begin{figure}[t]
\centering
\begin{tabular}{cc}
\includegraphics[scale=0.5]{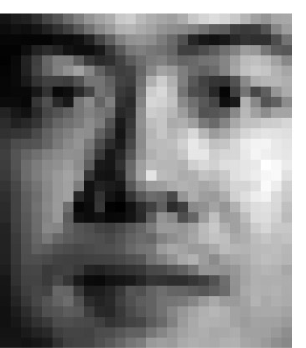} & \includegraphics[scale=0.5]{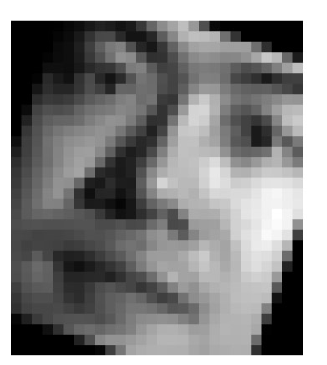}\\
(a) & (b)
\end{tabular}
\caption{An example of rotated test images. (a)~Original image and (b)~the image rotated by 20 degrees clockwise.}
\label{fig::rotation}
\end{figure}

We test the proposed algorithm(s) on popular face databases. Experiments performed in \cite{Wright_2009} reveal the robustness of the  approach to distortions, under the assumption that the test images are well-calibrated. As a first result, we show in Section \ref{section::calibrated} that our proposed approach produces equally competitive results on calibrated test images (with no registration errors) from the Extended Yale B database \cite{GeBeKr01}. Subsequently, via experiments in Section \ref{section::regnerr}, we establish the robustness of our approach to registration errors and a variety of other distortions. The ability to reject invalid images is tested in Section \ref{section::outreject}. Finally, we discuss different flavors of classifier fusion (to combine the local recognition decisions) in Section \ref{section::classfusion}. MATLAB code corresponding to all the experiments and algorithms reported in this paper is available at: {\tt http://signal.ee.psu.edu/FaceRec-LSGM.htm}.

\begin{figure}[t]
\centering
\includegraphics[scale=0.4]{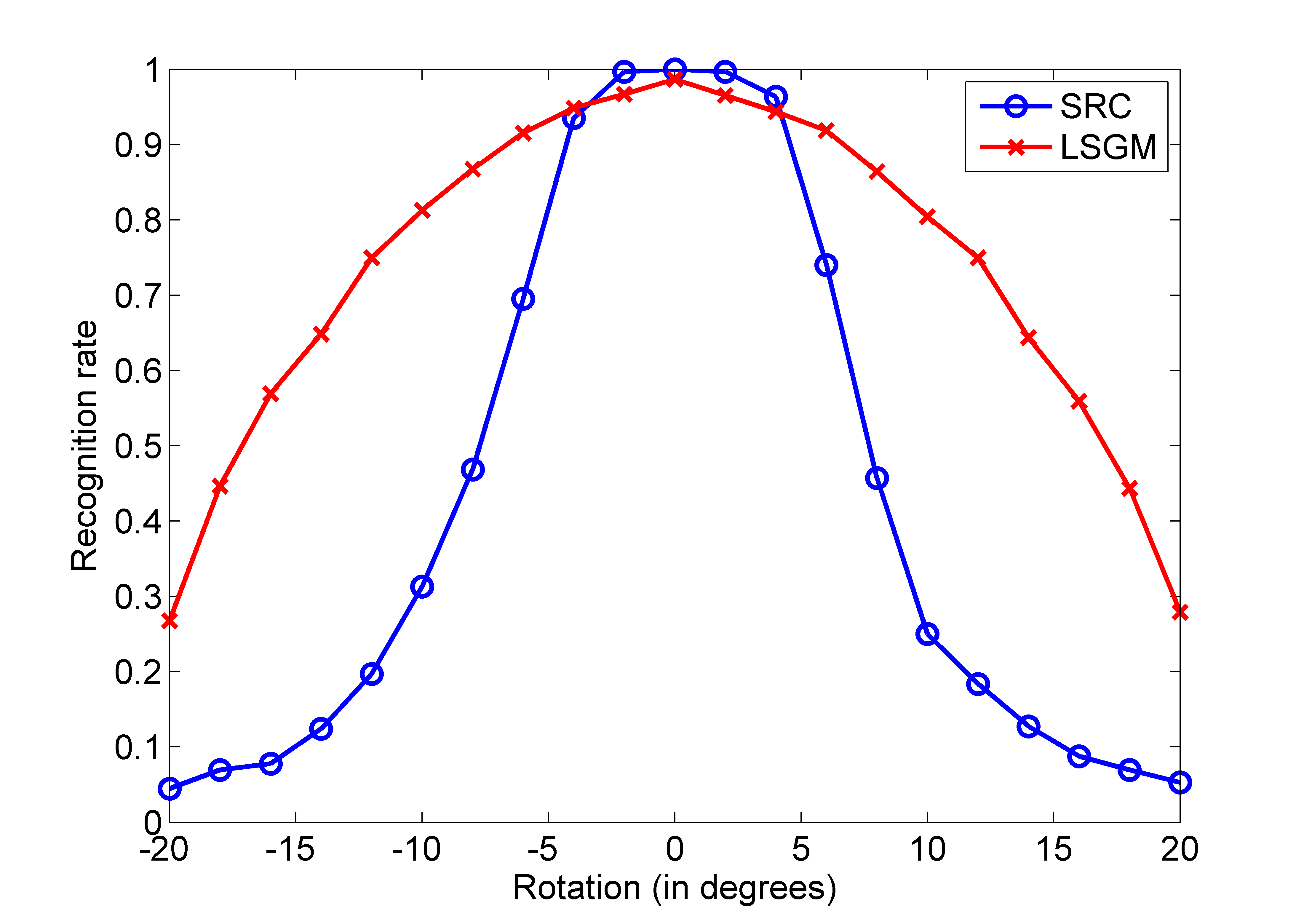}
\caption{Recognition rate for rotated test images (Section \ref{section::regnerr}).}
\label{fig::rotation_results}
\end{figure}

\begin{figure}[t]
\centering
\begin{tabular}{cc}
\includegraphics[scale=0.5]{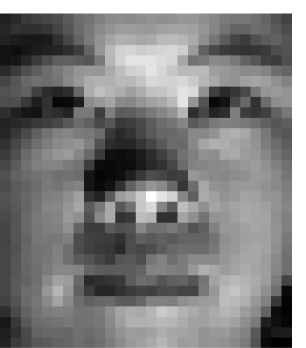} & \includegraphics[scale=0.5]{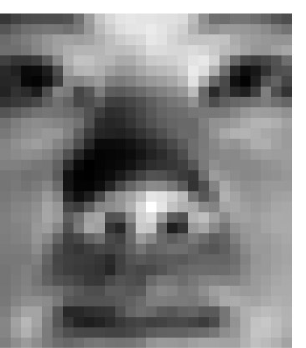}\\
(a) & (b)
\end{tabular}
\caption{An example of scaled test images. (a)~Original image and (b)~the image scaled by 1.313 vertically and 1.357 horizontally.}
\label{fig::zoom}
\end{figure}

\subsection{Calibrated Test Images: No Alignment Errors}
\label{section::calibrated}
For this experiment we use the Extended Yale B database, which consists of 2414 perfectly-aligned frontal face images of size $192\times 168$ of 38 individuals, 64 images per individual, under various conditions of illumination. In our experiments, for each subject we randomly choose 32 images in Subsets 1 and 2, which were taken under less extreme lighting conditions, as the training data. The remaining images are used as test data. All training and test samples are downsampled to size $32\times 28$.

In the following experiments, our face recognition algorithm comprises the extraction of local sparse features along with graphical model decisions (as described in Section \ref{section::contrib2}, part 2) which we term as Local-Sparse-GM abbreviated to LSGM. We compare our LSGM technique against five popular face recognition algorithms: (i) sparse representation-based classification (SRC) \cite{Wright_2009}, (ii) Eigenfaces \cite{Turk_1991} as features with nearest subspace \cite{Ho03} classifier (Eigen-NS), (iii) Eigenfaces with support vector machine \cite{Vapnik} classifier (Eigen-SVM), (iv) Fisherfaces \cite{Belhumeur_1997} as features with nearest subspace classifier (Fisher-NS), and (v) Fisherfaces with SVM classifier (Fisher-SVM). Overall recognition rates - ratio of the total number of correctly classified images to the total number of test images, expressed as a percentage - are reported in Table \ref{table::calibrated}. The results reveal that the choice of local sparse features over global features does not significantly affect the overall recognition performance in the scenario of no registration errors.

\begin{table}[t]
\begin{minipage}{\columnwidth}\centering%
\caption{Recognition rate (in percentage) for scaled test images using SRC \cite{Wright_2009} under various scaling factors (SF).}
\label{table::zoom_global}
\begin{tabular}{| l |l|l|l|l|l|l|}
\hline
SF & 1 & 1.071  &  1.143 &   1.214  &  1.286  &  1.357\\
\hline
    1     & 100    &100   &  98.0 &  88.2 &  76.5 &  58.8\\
\hline
    1.063 &  99.7  & 96.5 &  86.1 &  68.5 &  50.3 &  37.6\\
\hline
    1.125 &  83.8  & 70.2 &  49.8 &  33.6 &  26.2 &  17.9\\
\hline
    1.188 &  54.5  & 43.7 &  26.8 &  20.0 &  18.0 &  12.6\\
\hline
    1.25  &  36.1  & 27.2 &  20.9 &  16.6 &  12.3 &  11.3\\
\hline
    1.313 &  31.5  & 24.3 &  16.7 &  13.9 &  10.6 &   9.8\\
\hline
\end{tabular}
\end{minipage}
\end{table}
\begin{table}[t]
\begin{minipage}{\columnwidth}\centering%
\caption{Recognition rate (in percentage) for scaled test images using proposed block-based approach under various SF.}
\label{table::zoom_local}
\begin{tabular}{| l |l|l|l|l|l|l|}
\hline
SF & 1 & 1.071  &  1.143 &   1.214  &  1.286  &  1.357\\
\hline
    1     &  98.8   &  98.2 &  98.5 &  97.5  & 97.5  & 97.2\\
\hline
    1.063 &  97.5 &  96.7 &  96.0 &  96.0  & 93.5  & 93.4\\
\hline
    1.125 &  97.4   &  96.5 &  96.2 &  95.2  & 93.2  & 91.1\\
\hline
    1.188 &  94.9   &  92.9   &  91.6 &  89.4  & 87.1  & 83.3\\
\hline
    1.25  &  94.9 &  93.0 &  92.2   &  87.9    & 82.0  & 77.8\\
\hline
    1.313 &  90.7 &  90.4   &  84.1   &  81.0  & 75.5    & 64.2\\
\hline
\end{tabular}
\end{minipage}%
\end{table}

\subsection{Recognition Under Distortions and Misalignment}
\label{section::regnerr}
\subsubsection{Presence of registration errors}
\label{section::reg_rot_scale}
The primary motivation for our contribution in this paper is to achieve robust recognition under misalignment of test images. We create distorted test images in several ways and keep the training images unchanged, again using images from the Extended Yale B database. Robustness to image translation is ensured by simply choosing an appropriate search region for each block such that the corresponding blocks in the training images are included in the dictionary.

Next, we show experimental results for test images under rotation and scaling operations. In the first set of experiments, the test images are randomly rotated by an angle between -20 and 20 degrees, as illustrated by the example in Fig.~\ref{fig::rotation}. We compare the SRC approach with the proposed LSGM framework. Fig.~\ref{fig::rotation_results} shows the recognition rate ($y$-axis) for each rotation degree ($x$-axis). We see that when the misalignment becomes more severe, the LSGM algorithm outperforms the SRC approach by a significant margin.

For the second set of experiments, the test images are stretched in both directions by scaling factors up to 1.313 vertically and 1.357 horizontally. An example of an aligned image in the database and its distorted version to be tested are shown in Fig.~\ref{fig::zoom}. Tables~\ref{table::zoom_global} and~\ref{table::zoom_local} show the percentage of correct identification with various scaling factors. The first row and the first column in the tables indicate the scaling factors in the horizontal and vertical directions respectively. We again see that when there are large registration errors, the block-based algorithm leads to a better identification performance than the original algorithm.
We observe similar behaviors when the scaling factors are in the range of 0.8 to 1 (that is, the test image is shrunk comparing to the training images in the dictionary).

\begin{figure*}[t]
\centering
\subfigure[]{\includegraphics[scale=0.4]{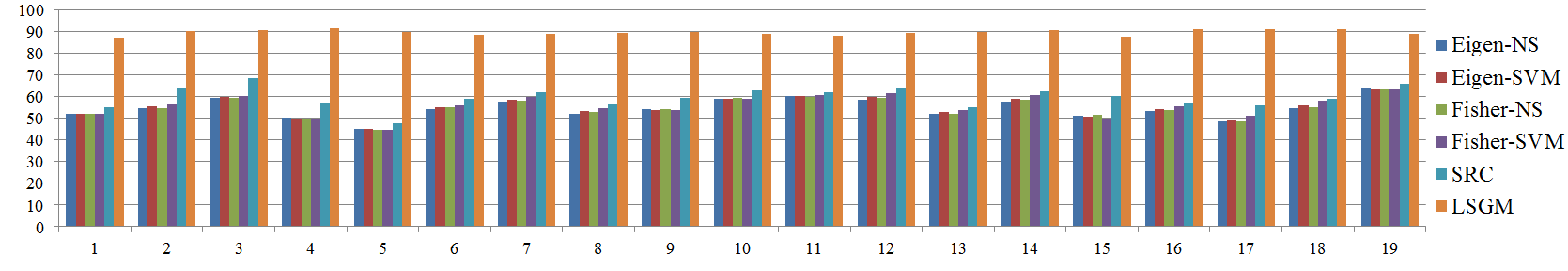}}
\subfigure[]{\includegraphics[scale=0.4]{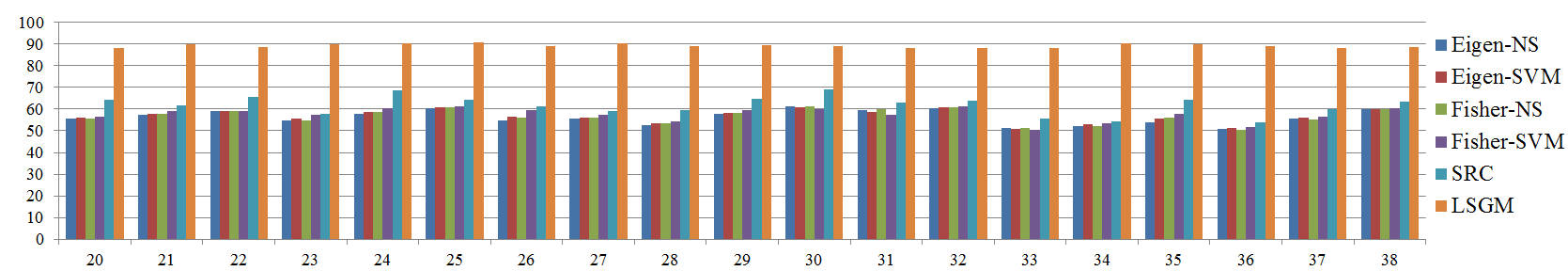}}
\caption{Face-specific recognition rates using the Extended Yale B database, with registration errors introduced in test images. (a) Results shown for faces numbered 1 through 19. (b) Results shown for faces numbered 20 through 38.}
\label{fig::distort_reg}
\end{figure*}

We now compare the performance of our LSGM approach with five other algorithms: SRC, Eigen-NS, Eigen-SVM, Fisher-NS and Fisher-SVM, for the particular scenario where the test images have been scaled by a horizontal factor of 1.214 and a vertical factor of 1.063. The per-face recognition rates are displayed for each approach in Fig. \ref{fig::distort_reg}, and the overall recognition rates are shown in Table \ref{table::regnerr}.

\begin{table}[t]
\begin{center}
\caption{Overall recognition rate (as a percentage) for the scenario of scaling by horizontal and vertical factors of 1.214 and 1.063 respectively.}
\begin{tabular}{|c|c|c|c|c|}
  \hline
  Method & Recognition rate ($\%$)\\
  \hline
  LSGM & 89.4 \\
  SRC & 60.8 \\
  Eigen-NS & 55.5 \\
  Eigen-SVM & 56.7\\
  Fisher-NS & 54.1\\
  Fisher-SVM & 57.1 \\
  \hline
\end{tabular}
\label{table::regnerr}
\end{center}
\end{table}

\begin{table}[t]
\begin{center}
\caption{Overall recognition rate (as a percentage) under registration errors, for images obtained from \cite{GTechDatabase}.}
\begin{tabular}{|c|c|c|c|c|}
  \hline
  Method & Recognition rate ($\%$)\\
  \hline
  LSGM & 87.6 \\
  SRC & 61.3 \\
  Eigen-NS & 47.4 \\
  Eigen-SVM & 50.5\\
  Fisher-NS & 45.3\\
  Fisher-SVM & 51.8 \\
  \hline
\end{tabular}
\label{table::regerr_gtech}
\end{center}
\end{table}

\begin{table}[t]
\begin{center}
\caption{Overall recognition rate (as a percentage) for the scenario where test images are scaled and subjected to random pixel corruption (Section \ref{section::pixcorr}).}
\begin{tabular}{|c|c|c|c|c|}
  \hline
  Method & Recognition rate ($\%$)\\
  \hline
  LSGM & 96.3 \\
  SRC & 93.2 \\
  Eigen-NS & 54.3 \\
  Eigen-SVM & 58.5\\
  Fisher-NS & 56.2 \\
  Fisher-SVM & 59.9 \\
  \hline
\end{tabular}
\label{table::pixcorr}
\end{center}
\end{table}

Next, we repeat the experiment on the Georgia Tech face database \cite{GTechDatabase}, wherein the test face captures are naturally frontal and/or tilted with different facial expressions, lighting conditions and scale. This database contains 15 faces each of 50 different individuals. For convenience, we restrict the data set to 38 classes of faces (chosen with no particular preference). We use five images from each class for training and the rest for testing. Here too, we provide a comparison of the per-face recognition rates for the LSGM method, and compare it with the five other approaches. The overall rates in Table \ref{table::regerr_gtech} confirm once again the robustness of the LSGM to misalignments in test images.

\subsubsection{Recognition despite random pixel corruption}
\label{section::pixcorr}
We return to the Extended Yale database for this experiment, where we randomly corrupt 50\% of the image pixels in each test image. In addition, each test image is scaled by a horizontal factor of 1.071 and a vertical factor of 1.063. Local sparse features are extracted using the robust form of the $\ell_1$-minimization similar to the approach in \cite{Wright_2009}. The overall recognition rates are shown in Table \ref{table::pixcorr}. These results reveal that under the mild scaling distortion scenario, our LSGM approach retains the robustness characteristic of the global sparsity approach (SRC), while the other competitive algorithms suffer drastic degradation in performance.

\subsubsection{Recognition despite disguise}
\label{section::disguise}
We test the robustness of our proposed LSGM approach to disguise (representative of real-life scenarios) using the AR Face Database \cite{ARDatabase}. We choose a subset of the database containing 50 male and 50 female subjects chosen randomly. For training, we consider 8 clean (with no occlusions) images each per subject. These images may however capture different facial expressions. Faces with two different types of disguise are used for testing purposes: subjects wearing sunglasses and subjects partially covering their face with a scarf. Accordingly, we present two sets of results. In each scenario, we use 6 images per subject for testing, leading to a total of 600 test images each for sunglasses and scarves. Consistent with our other experiments, we also introduce mild misalignment in the test images, in the form of scaling by horizontal and vertical factors of 1.071 and 1.063 respectively.

To enable robustness against disguise, in \cite{Wright_2009} the authors also suggest block partitioning to improve the results, by aggregating results from individual blocks using voting. It is useful to point out two key differences between this strategy and our proposed approach: (i) we use an adaptive local block-based model to build the training dictionary to incorporate robustness to misalignment, and (ii) we use a principled classification framework using graphical models to combine results from the individual blocks rather than simple voting.

The results of our proposed approach (using three representative local regions) are compared with five other competitive approaches in Table \ref{table::disguise}. The LSGM and SRC approaches significantly outperform the other methods. Further, the improvements of LSGM over SRC reveal the benefits of the graphical model framework for classification over the voting scheme. For additional improvements in recognition rate, we can use a larger number of local spatial blocks.

\begin{table}[t]
\begin{center}
\caption{Overall recognition rate (as a percentage) for the scenario where test images are scaled and subjects wear disguise (Section \ref{section::disguise}).}
\begin{tabular}{|c|c|c|}
  \hline
  Method & Recognition rate ($\%$) & Recognition rate ($\%$) \\
  & Sunglasses & Scarves \\
  \hline
  LSGM & 96.0 & 92.9\\
  SRC & 93.5 & 90.1\\
  Eigen-NS & 47.2 & 29.6 \\
  Eigen-SVM & 53.5 & 34.5 \\
  Fisher-NS & 57.9 & 41.7\\
  Fisher-SVM & 61.7 & 43.6\\
  \hline
\end{tabular}
\label{table::disguise}
\end{center}
\end{table}

\begin{figure}[thp]
\centering
\includegraphics[scale=0.45]{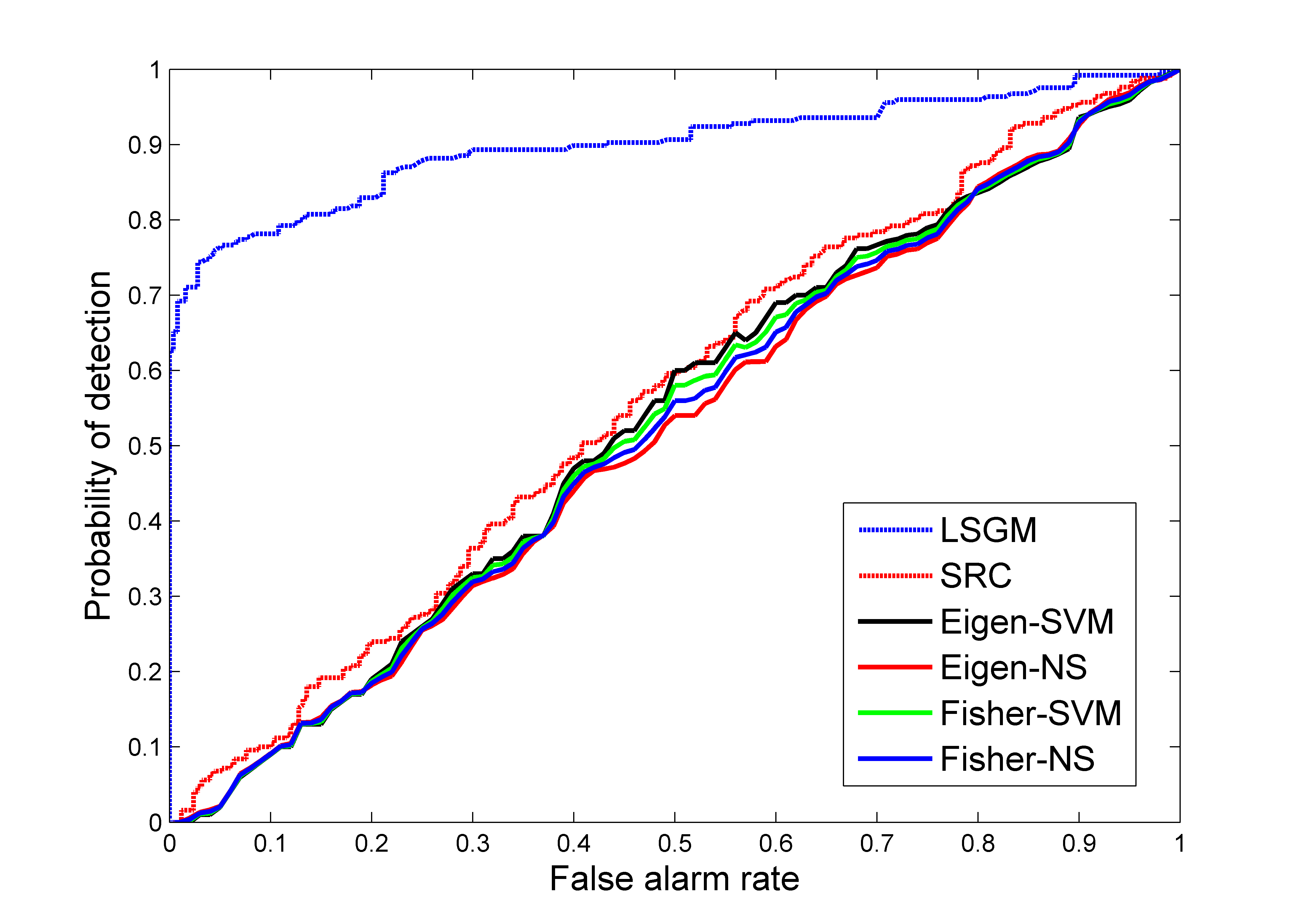}
\vspace{-2mm}
\caption{ROC curves for outlier rejection (Section \ref{section::outreject}).}
\label{fig::ROC_rotate5}
\vspace{-5mm}
\end{figure}

\subsection{Outlier Rejection}
\label{section::outreject}
In this experiment, samples from 19 of the 38 classes in the Yale database are included in the training set, and faces from the other 19 classes are considered outliers. For training, 15 samples per class from Subsets 1 and 2 are used~($19\times15 = 285$ samples in total), while 500 samples are randomly chosen for testing, among which 250 are inliers and the other 250 are outliers. All test samples are rotated by five degrees.

The five different competing approaches are compared with our proposed LSGM method.
For the LSGM approach, we use a minimum threshold $\delta$ on the quantity described in \eqref{eq:decision_rule}. If the maximum value of the log-likelihood ratio does not exceed $\delta$, the corresponding test sample is labeled an outlier. In the SRC approach, the Sparsity Concentration Index~\eqref{eq::sci} is used as the criterion for outlier rejection. For the other approaches under comparison which use the nearest subspace and SVM classifiers, reconstruction residuals are compared to a threshold to decide outlier rejection. The receiver operating characteristic~(ROC) curves for all the approaches are shown in Fig.~\ref{fig::ROC_rotate5}, where the probability of detection is the ratio between the number of detected inliers and the total number of inliers, and the false alarm rate is computed by the number of outliers which are detected as inliers divided by the total number of outliers. Under scaling distortion, we see that LSGM offers the best performance, while some of the approaches are actually worse than random guessing.
\begin{figure*}[htp]
\centering
\subfigure[]{\includegraphics[scale=0.4]{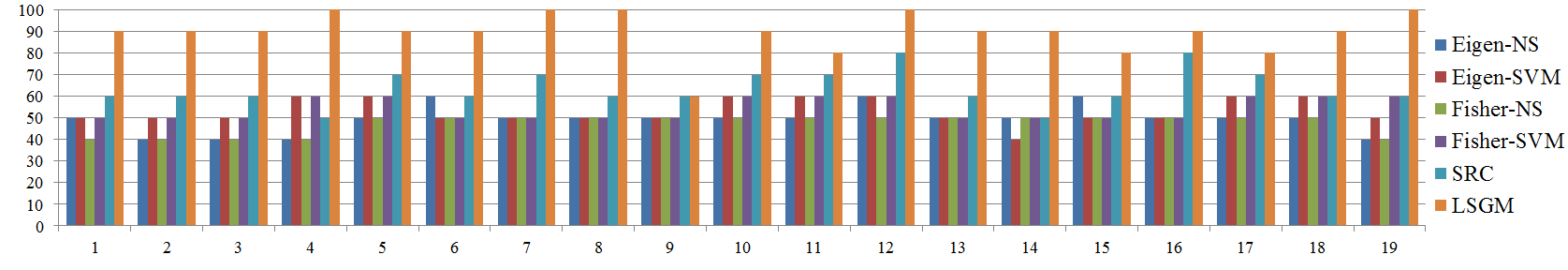}}
\subfigure[]{\includegraphics[scale=0.4]{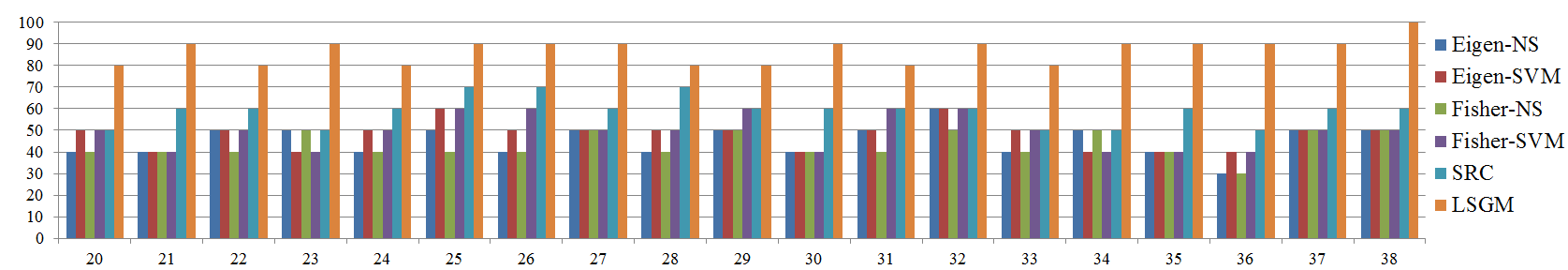}}
\caption{Face-specific recognition rates using the Georgia Tech face database, with registration errors introduced in test images. (a) Results shown for faces numbered 1 through 19. (b) Results shown for faces numbered 20 through 38.}
\label{fig::regerr_gtech}
\end{figure*}

\subsection{Classifier Fusion: Variants of Proposed Method}
\label{section::classfusion}
We now compare the performance of the different proposed ways of combining the local classifier decisions from Section \ref{section::contrib2}: (i) majority voting (Voting), (ii) heuristic maximum likelihood (ML)-type fusion using reconstruction residuals (LHML), and (iii) the discriminative graphical model framework (LSGM). The images are taken from the Extended Yale B Database. We introduce mild misalignment in the test images in the form of scaling by a horizontal factor of 1.214 and a vertical factor of 1.063. We use 15 training samples per class, and a total of 1844 samples for testing.

\begin{figure*}[htp]
\centering
\subfigure[]{\includegraphics[scale=0.4]{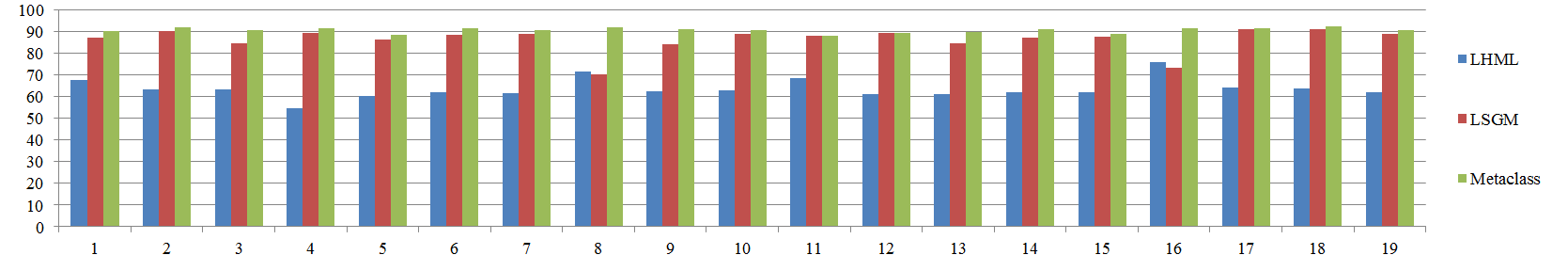}}
\subfigure[]{\includegraphics[scale=0.4]{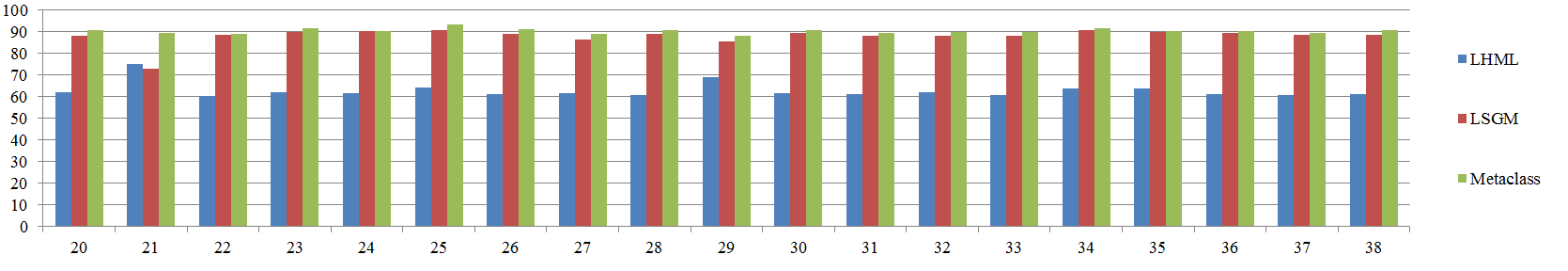}}
\caption{Meta-classification: Face-specific recognition rates using the Extended Yale B face database, with scaling registration errors introduced in test images. (a) Results shown for faces numbered 1 through 19. (b) Results shown for faces numbered 20 through 38.}
\label{fig::metaclass}
\end{figure*}

Although the LSGM approach has superior overall recognition performance in comparison to the Voting and LHML techniques, we see from Fig. \ref{fig::metaclass} that for some of the classes, the LHML approach in fact offers slightly better recognition rates. So, we propose a \emph{principled} meta-classification framework to further exploit these complementary benefits offered by the individual classifiers. From each type of classifier, we obtain ``soft'' outputs, that estimate the posterior probability of a face belonging to a particular class. These soft outputs may also be interpreted as indicating the degree of confidence in the decision. These outputs may then be treated as meta-feature vectors to be fed into a support vector machine for meta-classification. We train the SVM using the soft outputs obtained from the training samples. A radial basis function (RBF) kernel is used in the SVM.

For perfectly calibrated test images, voting presents a computationally simple way of benefitting from the classification results from individual local blocks. However, in the presence of registration errors, voting performs poorly, leading to reduced overall performance of the meta-classifier. So, we present results using only two classifiers: LHML and LSGM. The per-class rates for the individual schemes as well as the meta-classifier are presented in Fig. \ref{fig::metaclass}. Meta-classification shows that the complementary benefits of different classifiers can be mined to improve recognition performance.

\subsection{Influence of number of local blocks on recognition performance}
So far, we have used three preceptively meaningful local blocks for the LSGM approach, while proposing the use of 42 uniformly sampled blocks of size $8\times 8$ for the LHML method. Unsurprisingly, the presence of more local blocks can improve recognition by offering more robustness to distortions. So, in this section, we evaluate the performance of our proposed algorithms as a function of number of blocks. Specifically, we use 3, 5, 8, 12, 20, 30 or 42 blocks in different experiments. For the case of 5 blocks, we pick the five (perceptually most meaningful) regions to be the block of two eyes, nose, mouth, and the two eyes taken individually. For larger number of blocks, the blocks are chosen uniformly from the entire image and the size of the blocks is either $8\times 12$ or $8\times 8$.

\begin{figure}[t]
\centering
\includegraphics[scale=0.4]{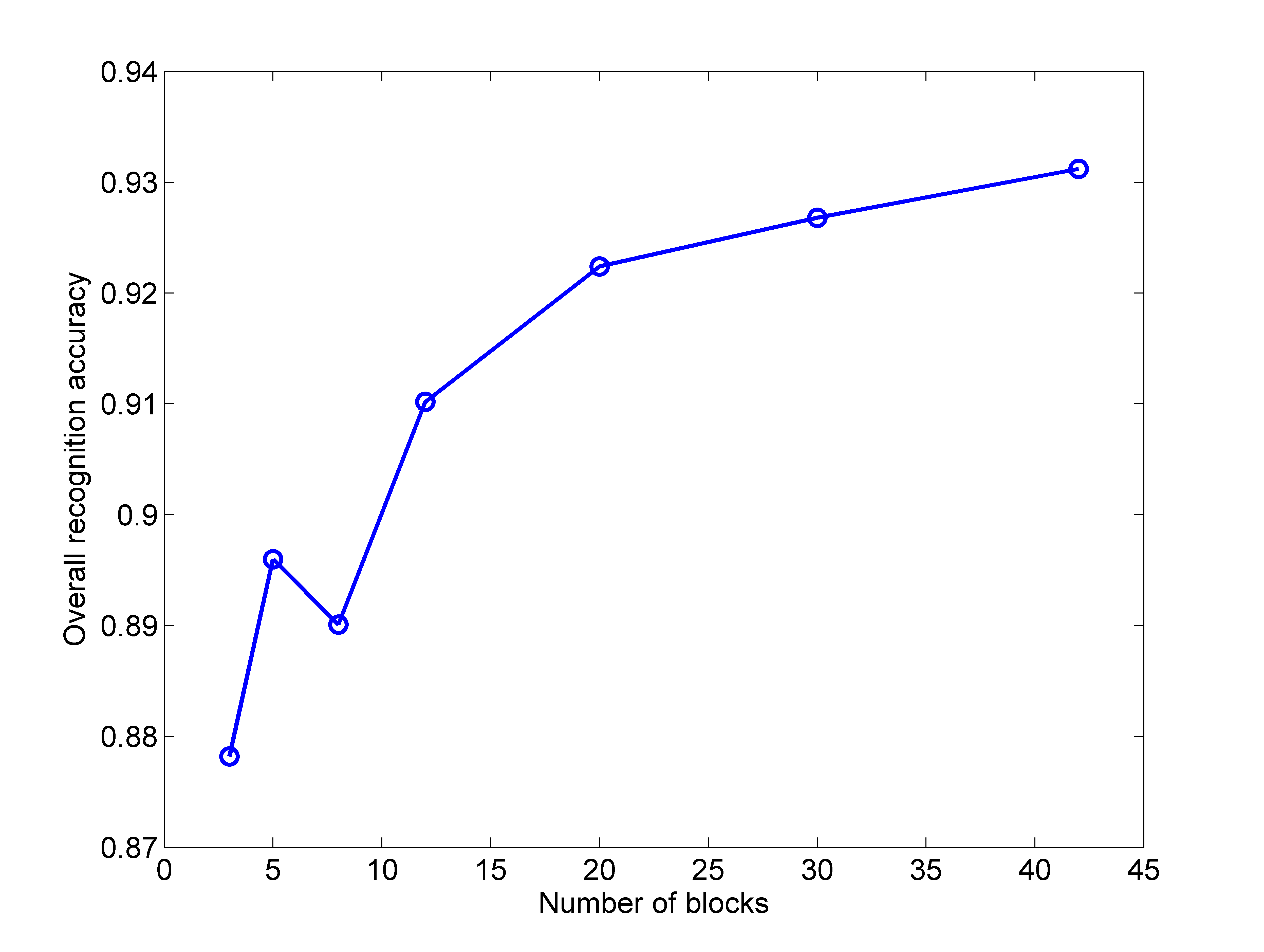}
\caption{Recognition performance of LSGM as a function of number of local blocks. Experiments are performed on the Georgia Tech database.}
\label{fig::gtech_blocksize}
\end{figure}

\begin{figure}[htp]
\centering
\includegraphics[scale=0.45]{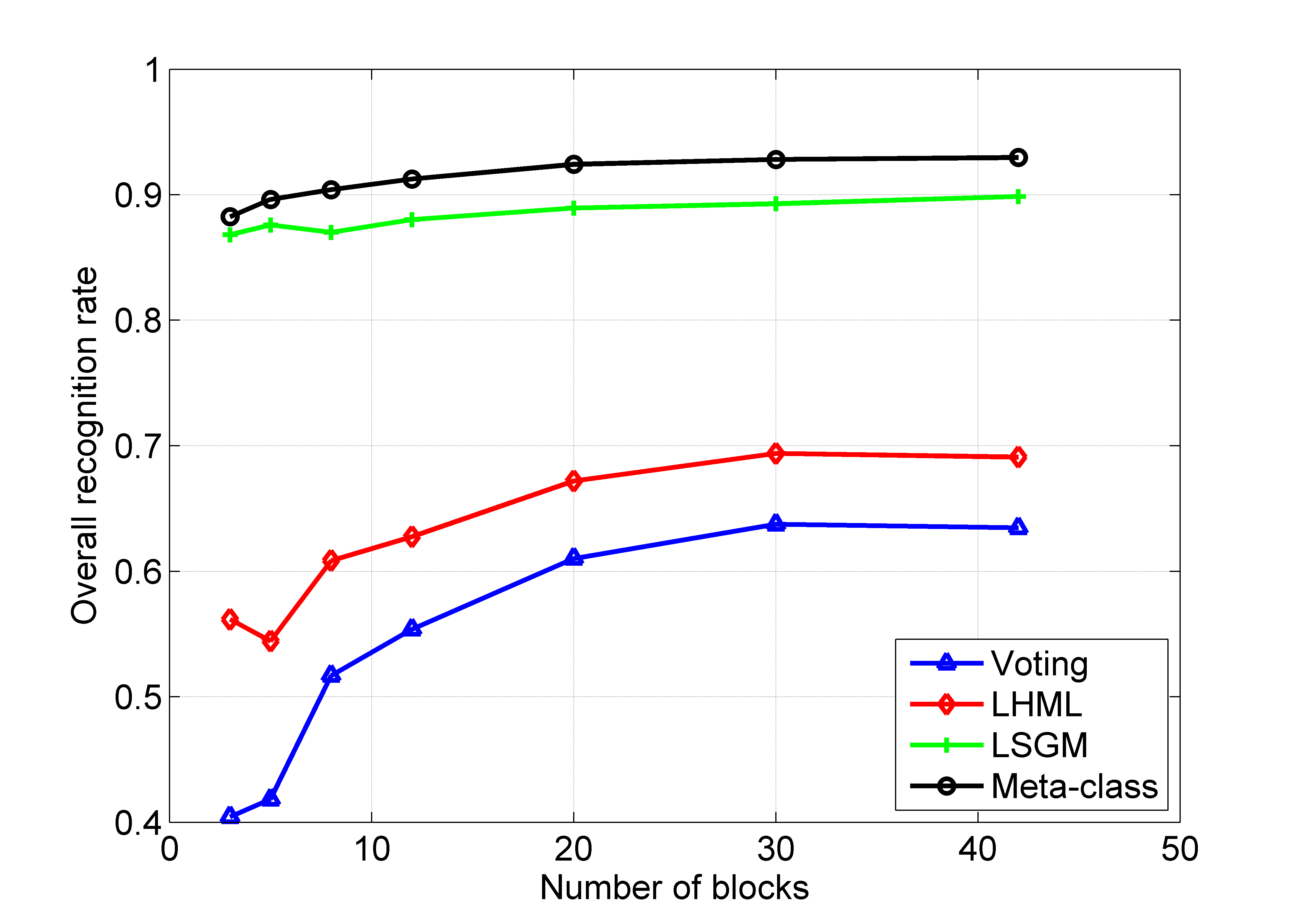}
\caption{Recognition performance of proposed classifiers and meta-classifier, as a function of number of local blocks.}
\label{fig::metaclass_blocksize}
\end{figure}

We choose two specific experiments to illustrate the dependence on number of blocks. First, we consider images from the Georgia Tech database, where the test images are naturally misaligned (Section \ref{section::reg_rot_scale}). The performance of the LSGM approach is shown in Fig. \ref{fig::gtech_blocksize}. There is a dip in recognition performance for the case of 8 blocks compared to the case of 5 blocks, since the 8 blocks are chosen uniformly from the image and need not necessarily carry perceptual meaning, while the 5 blocks are chosen in a particular meaningful manner. However, with the increase in the number of blocks, the particular choice of blocks seemingly becomes less relevant.

For the second experiment, we consider the meta-classification scenario described earlier in this section. The resulting plot is showed in Fig. \ref{fig::metaclass_blocksize}. The voting approach performs very poorly in comparison with the LHML and LSGM approaches. As expected, the meta-classifier improves upon the performance of all the methods. More significantly, Fig. \ref{fig::metaclass_blocksize} reveals that the LSGM approach is less sensitive to variations in the number of blocks and particular choice of blocks, while the performance of other proposed local approaches is contingent on the availability of sufficient number of local blocks.

\section{Conclusion}
\label{section::conclusion}

We developed a local block-based sparsity model to realize a practical face recognition algorithm which exhibits robustness to alignment errors and a host of distortions such as noise, occlusion, disguise and illumination changes. Unlike other competing techniques, no explicit registration step is required - which makes our approach computationally simpler. Inspired by human perception, our sparse local features are extracted via projections onto adaptive dictionaries built from informative regions of the face image such as eyes, nose and mouth. Instead of using class specific reconstruction error (which does not capture inter-class variation), we present a probabilistic graphical model framework to explicitly capture the conditional correlations between these sets of local features. Experiments on benchmark face databases and comparisons against state-of-the-art face recognition techniques under numerous practical testing environments reveal the merits of our proposal.

\bibliographystyle{IEEEtran}
\bibliography{IEEEabrv,ICIP10bib}

\begin{thebibliography}{10}
\providecommand{\url}[1]{#1}
\csname url@samestyle\endcsname
\providecommand{\newblock}{\relax}
\providecommand{\bibinfo}[2]{#2}
\providecommand{\BIBentrySTDinterwordspacing}{\spaceskip=0pt\relax}
\providecommand{\BIBentryALTinterwordstretchfactor}{4}
\providecommand{\BIBentryALTinterwordspacing}{\spaceskip=\fontdimen2\font plus
\BIBentryALTinterwordstretchfactor\fontdimen3\font minus
  \fontdimen4\font\relax}
\providecommand{\BIBforeignlanguage}[2]{{%
\expandafter\ifx\csname l@#1\endcsname\relax
\typeout{** WARNING: IEEEtran.bst: No hyphenation pattern has been}%
\typeout{** loaded for the language `#1'. Using the pattern for}%
\typeout{** the default language instead.}%
\else
\language=\csname l@#1\endcsname
\fi
#2}}
\providecommand{\BIBdecl}{\relax}
\BIBdecl

\bibitem{Zhao_2003}
W.~Zhao, R.~Chellappa, P.~J. Phillips, and A.~Rosenfeld, ``Face recognition: A
  literature survey,'' \emph{ACM Computing Surveys}, vol.~35, no.~4, pp.
  399--458, Dec. 2003.

\bibitem{Sirovich_1987}
L.~Sirovich and M.~Kirby, ``Low-dimensional procedure for the characterization
  of human faces,'' \emph{J. Optical Soc. of Am. A}, vol.~4, no.~3, pp.
  519--524, Mar. 1987.

\bibitem{Turk_1991}
M.~Turk and A.~Pentland, ``Eigenfaces for recognition,'' \emph{J. Cogn.
  Neurosci.}, vol.~3, no.~1, pp. 71--86, Winter 1991.

\bibitem{Zou07}
J.~Zou, Q.~Ji, and G.~Nagy, ``A comparative study of local matching approach
  for face recognition,'' \emph{{IEEE} Trans. Image Process.}, vol.~16, no.~10,
  pp. 2617--2628, Oct. 2007.

\bibitem{Shashua_1992}
A.~Shashua, ``{Geometry and photometry in 3D visual recognition},'' Ph.D.
  dissertation, MIT, 1992.

\bibitem{Belhumeur_1997}
P.~N. Belhumeur, J.~P. Hespanha, and D.~J. Kriegman, ``Eigenfaces vs.
  fisherfaces: Recognition using class specific linear projection,''
  \emph{{IEEE} Trans. Pattern Anal. Mach. Intell.}, vol.~19, no.~7, pp.
  711--720, Jul. 1997.

\bibitem{Liu01}
C.~Liu and H.~Wechsler, ``A shape- and texture-based enhanced fisher classifier
  for face recognition,'' \emph{{IEEE} Trans. Image Process.}, vol.~10, no.~4,
  pp. 598--608, Apr. 2001.

\bibitem{Basri_2003}
R.~Basri and D.~W. Jacobs, ``Lambertian reflectance and linear subspaces,''
  \emph{{IEEE} Trans. Pattern Anal. Mach. Intell.}, vol.~25, no.~2, pp.
  218--233, Feb. 2003.

\bibitem{Wright_2009}
J.~Wright, A.~Y. Yang, A.~Ganesh, S.~Sastry, and Y.~Ma, ``Robust face
  recognition via sparse representation,'' \emph{{IEEE} Trans. Pattern Anal.
  Mach. Intell.}, vol.~31, no.~2, pp. 210--227, Feb. 2009.

\bibitem{Pillai_2011}
J.~K. Pillai, V.~M. Patel, R.~Chellappa, and N.~Ratha, ``Secure and robust iris
  recognition using sparse representations and random projections,''
  \emph{{IEEE} Trans. Pattern Anal. Mach. Intell.}, vol.~33, no.~9, pp.
  1877--1893, Sep. 2011.

\bibitem{Hang_2009}
X.~Hang and F.-X. Wu, ``Sparse representation for classification of tumors
  using gene expression data,'' \emph{Journal of Biomedicine and
  Biotechnology}, vol. 2009, 2009, doi:10.1155/2009/403689.

\bibitem{Huang_2008}
J.~Huang, X.~Huang, and D.~Metaxas, ``Simultaneous image transformation and
  sparse representation recovery,'' in \emph{Proc. of IEEE Conf. Comput. Vision
  Pattern Recognition}, Anchorage, AK, Jun. 2008, pp. 1--8.

\bibitem{Wagner_2009}
A.~Wagner, J.~Wright, A.~Ganesh, Z.~Zhou, and Y.~Ma, ``Towards a practical face
  recognition system: Robust registration and illumination by sparse
  representation,'' in \emph{Proc. of IEEE Conf. Comput. Vision Pattern
  Recognition}, Miami, FL, Jun. 2009, pp. 597--604.

\bibitem{Wagner:robust_face}
A.~Wagner, J.~Wright, A.~Ganesh, Z.~Zhou, H.~Mobahi, and Y.~Ma, ``Towards a
  practical face recognition system: {R}obust alignment and illumination by
  sparse representation,'' \emph{{IEEE} Trans. Pattern Anal. Mach. Intell.}, to
  appear.

\bibitem{Chen_2009}
Y.~Chen, T.~T. Do, and T.~D. Tran, ``Robust face recognition using locally
  adaptive sparse representation,'' in \emph{Proc. {IEEE} Intl. Conf. Image
  Processing}, Hong Kong, Sep. 2011, pp. 1657--1660.

\bibitem{Kittler98}
J.~Kittler, M.~Hatef, R.~P.~W. Duin, and J.~Matas, ``On combining
  classifiers,'' \emph{{IEEE} Trans. Pattern Anal. Mach. Intell.}, vol.~20,
  no.~3, pp. 226--239, Mar. 1998.

\bibitem{faceRec:Srinivas11}
U.~Srinivas, V.~Monga, Y.~Chen, and T.~D. Tran, ``Sparsity-based face
  recognition using discriminative graphical models,'' in \emph{Proc. {IEEE}
  Asilomar Conf. on Signals, Systems and Computers}, Pacific Grove, CA, Nov.
  2011.

\bibitem{Tan10}
V.~Y.~F. Tan, S.~Sanghavi, J.~W.~F. III, and A.~S. Willsky, ``Learning
  graphical models for hypothesis testing and classification,'' \emph{IEEE
  Trans. Signal Processing}, vol.~58, no.~11, pp. 5481--5495, Nov. 2010.

\bibitem{freund:Adaboost}
Y.~Freund and R.~E. Schapire, ``A short introduction to boosting,''
  \emph{Journal of Japanese Society for Artificial Intelligence}, vol.~14,
  no.~5, pp. 771--780, Sep. 1999.

\bibitem{CandesRob06}
E.~Cand\`{e}s, J.~Romberg, and T.~Tao, ``Robust uncertainty principles: Exact
  signal reconstruction from highly incomplete frequency information,''
  \emph{{IEEE} Trans. Inf. Theory}, vol.~52, no.~2, pp. 489--509, Feb. 2006.

\bibitem{Tropp_2005}
J.~Tropp and A.~Gilbert, ``Signal recovery from random measurements via
  orthogonal matching pursuit,'' \emph{{IEEE} Trans. Inf. Theory}, vol.~53,
  no.~12, pp. 4655--4666, Dec. 2007.

\bibitem{Dai_2009}
W.~Dai and O.~Milenkovic, ``Subspace pursuit for compressive sensing signal
  reconstruction,'' \emph{{IEEE} Trans. Inf. Theory}, vol.~55, no.~5, pp.
  2230--2249, May 2009.

\bibitem{Do_2008}
T.~T. Do, L.~Gan, N.~H. Nguyen, and T.~D. Tran, ``Sparsity adaptive matching
  pursuit algorithm for practical compressed sensing,'' in \emph{Proc. IEEE
  Asilomar Conf. on Signals, Systems, and Computers}, Pacific Grove, CA, Oct.
  2008, pp. 581--587.

\bibitem{Lauritzen96}
S.~L. Lauritzen, \emph{{Graphical Models}}.\hskip 1em plus 0.5em minus
  0.4em\relax Oxford University Press, NY, 1996.

\bibitem{wainwright:GMs}
M.~J. Wainwright and M.~I. Jordan, ``Graphical models, exponential families and
  variational inference,'' \emph{Foundations and Trends in Machine Learning},
  vol.~1, no. 1-2, pp. 1--305, 2008.

\bibitem{Chow68}
C.~K. Chow and C.~N. Liu, ``Approximating discrete probability distributions
  with dependence trees,'' \emph{{IEEE} Trans. Inf. Theory}, vol.~14, no.~3,
  pp. 462--467, Mar. 1968.

\bibitem{Do_2009}
T.~T. Do, Y.~Chen, D.~T. Nguyen, N.~H. Nguyen, L.~Gan, and T.~D. Tran,
  ``Distributed compressed video sensing,'' in \emph{Proc. {IEEE} Int. Conf.
  Image Process.}, Cairo, Egypt, Nov. 2009, pp. 1393--1396.

\bibitem{Elad_2006}
M.~Elad and M.~Aharon, ``Image denoising via sparse and redundant
  representations over learned dictionaries,'' \emph{{IEEE} Trans. Image
  Process.}, vol.~15, no.~12, pp. 3736--3745, Dec. 2006.

\bibitem{GeBeKr01}
A.~S. Georghiades, P.~N. Belhumeur, and D.~J. Kriegman, ``From few to many:
  Illumination cone models for face recognition under variable lighting and
  pose,'' \emph{{IEEE} Trans. Pattern Anal. Mach. Intell.}, vol.~23, no.~6, pp.
  643--660, Jun. 2001.

\bibitem{Friedman97}
N.~Friedman, D.~Geiger, and M.~Goldszmidt, ``Bayesian network classifiers,''
  \emph{Machine Learning}, vol.~29, pp. 131--163, Nov. 1997.

\bibitem{discGM:Srinivas11}
U.~Srinivas, V.~Monga, and R.~G. Raj, ``Automatic target recognition using
  discriminative graphical models,'' in \emph{Proc. {IEEE} Intl. Conf. Image
  Processing}, Brussels, Belgium, Sep. 2011, pp. 33--36.

\bibitem{Ho03}
J.~Ho, M.~Yang, J.~Lim, K.~Lee, and D.~Kriegman, ``Clustering appearances of
  objects under varying illumination conditions,'' in \emph{Proc. of IEEE Conf.
  Comput. Vision Pattern Recognition}, Madison, WI, Jun. 2003, pp. 11--18.

\bibitem{Vapnik}
V.~N. Vapnik, \emph{The nature of statistical learning theory}.\hskip 1em plus
  0.5em minus 0.4em\relax New York, USA: Springer, 1995.

\bibitem{GTechDatabase}
\BIBentryALTinterwordspacing
``{\em The Georgia Tech Face Database}.'' [Online]. Available:
  \url{http://www.anefian.com/research/face_reco.htm}
\BIBentrySTDinterwordspacing

\bibitem{ARDatabase}
A.~M. Martinez and R.~Benavente, ``{The AR face database},'' \emph{CVC Tech.
  Report}, no.~24, 1998.

\end{thebibliography}


\end{document}